\documentclass[lettersize,journal]{IEEEtran}
\usepackage{amsmath,amsfonts}
\usepackage{algorithmic}
\usepackage{algorithm}
\usepackage{array}
\usepackage[caption=false,font=normalsize,labelfont=sf,textfont=sf]{subfig}
\usepackage{textcomp}
\usepackage{stfloats}
\usepackage{url}
\usepackage{verbatim}
\usepackage{graphicx}
\usepackage{cite}
\usepackage{amsmath}
\usepackage{amssymb}
\usepackage{multirow}
\usepackage{array}
\usepackage{makecell}
\usepackage[table]{xcolor}
\usepackage[flushleft]{threeparttable}
\usepackage{bm}
\usepackage{subfig}
\usepackage{graphicx}
\usepackage{subfig}
\usepackage{caption}   
\usepackage{booktabs}
\begin{document}

\title{Smooth bounding box for object detection}

\author{Chao Zhou, Shenghua He, Fengchuang Xing, Han Yang, Qiangguo Yu
\thanks{Chao Zhou and Qiangguo Yu are with the School of Electronic Information, Huzhou College, Huzhou, 313000, China (e-mail: zhouchaodzkd@e.gzhu.edu.cn; yuqg@zjhzu.edu.cn).}
\thanks{Shenghua He is with the School of Electronic Science and Engineering, South China Normal University, Foshan, 528225, China (e-mail: acuicuiemail@163.com).}
\thanks{Fengchuang Xing is with the School of Physics and Information Engineering, Guangdong University of Education, Guangzhou, 510303, China (e-mail: xfchuang@e.gzhu.edu.cn).}
\thanks{Han Yang is an independent researcher, Lijiang, 674200, China (e-mail: 15812244636@163.com).}

\thanks{Shenghua He and Chao Zhou are equally contributed to this work.}
}



\maketitle

\begin{abstract}
Deep learning-based object detection models are highly dependent on the quality of annotated training data. However, large-scale datasets often contain inevitable label noise, with bounding box localization errors being a prevalent and impactful form in object detection. While label noise handling techniques are well-studied for image classification, their adaptation to the more complex problem of bounding box errors remains underexplored.
To address this, we propose a novel Smooth Bounding Box (SBB) strategy to mitigate the adverse effects of bounding box annotation inaccuracies. Our key insight is that annotation errors are often linked to annotator factors rather than specific image content, allowing for a statistical correction approach. Based on object size distribution within datasets, we implement the SBB strategy through two methods: 1) Global Scaling Correction (GSC), which applies a uniform scaling factor to estimate and correct a global average error; and 2) Interval Scaling Correction (ISC), which categorizes objects by size into intervals and estimates a dedicated correction factor for each, enabling more precise, fine-grained calibration. For efficient optimization of multiple parameters in ISC, we design a tailored model fine-tuning strategy. Furthermore, we introduce a Normalized Residual Fluctuation (NRF) metric to quantitatively assess the smoothing effect and training stability.
Extensive experiments on multiple benchmarks demonstrate that our method consistently enhances detection accuracy across various model architectures. Both GSC and ISC achieve significant performance improvements over strong baselines, including the latest YOLOv11n, without adding inference overhead. 
\end{abstract}


\begin{figure*}[tp]
\centering
\includegraphics[width=1\textwidth]{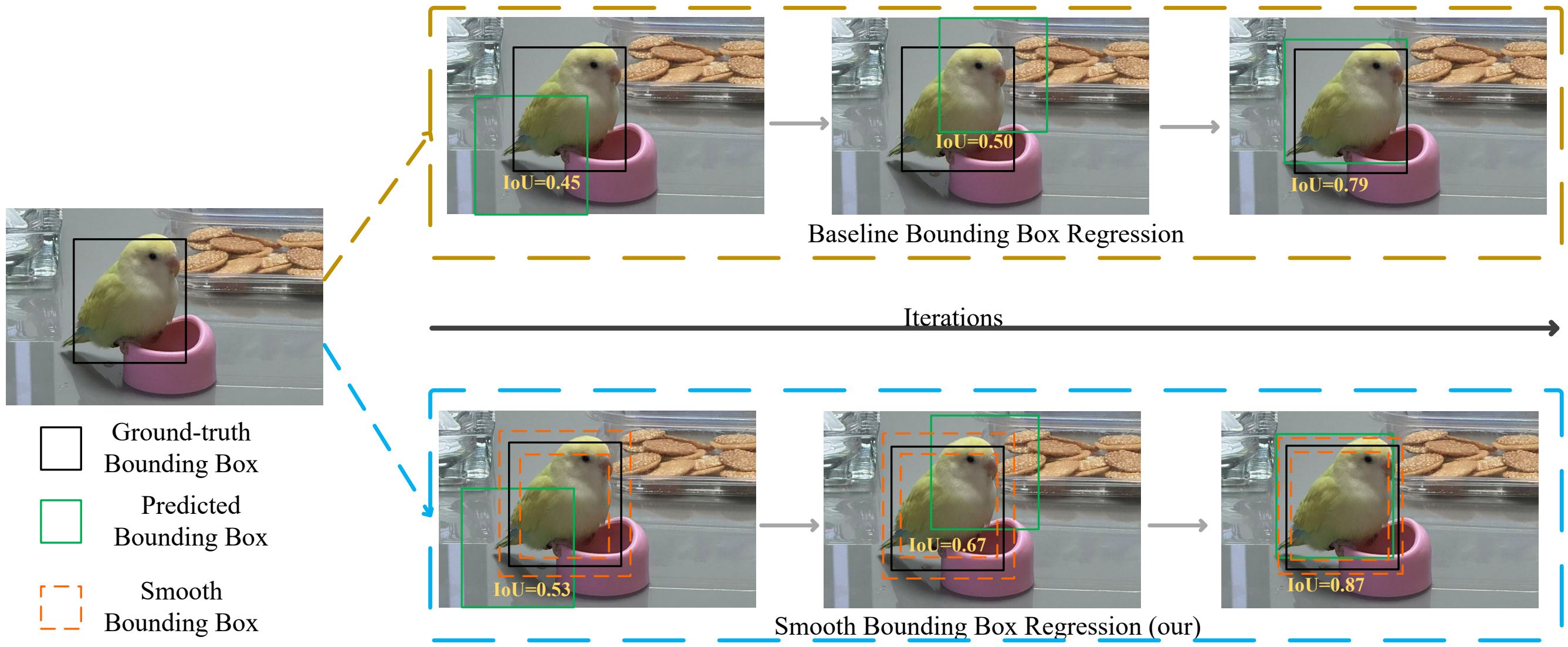}
\caption{Illustration of anchor box calibration during training. Uncertainty is introduced to the bounding box label through smooth expansion and contraction operations (indicated by orange boxes). For each predicted box, we calculate its IoU with the original ground truth and the two smoothed variants, then select the highest IoU value to calibrate the regression target.
}

\label{fig_smooth_lteration}

\end{figure*}

\section{Introduction}
\IEEEPARstart{N}{owadays, } deep neural networks (DNNs) have achieved great success in various computer vision tasks, such as object detection \cite{10133831,9693947, 10946136,10474116}. However, the performance of supervised deep learning methods heavily depend on the datasets. When the data volume is comparable, the quality of the dataset often becomes the decisive factor. Therefore, collecting large-scale, well-annotated datasets is crucial for the successful application of deep learning~\cite{krizhevsky2012imagenet}. The use of crowdsourcing or machine-generated labels has become a widely adopted practice~\cite{su2012crowdsourcing}, and these methods inevitably introduce noise. 
Compared to image classification datasets, label noise in object detection datasets exhibits more complex and diverse forms, primarily including class labeling errors, bounding box localization deviations, missing annotations, and false positive bounding boxes~\cite{ryoo2023universal}.

To reduce the influence of label errors in the field of image classification, various techniques have been developed. For example, Label Smoothing (LS)~\cite{muller2019does, szegedy2016rethinking} introduces uncertainty directly into the labels, while Learning with Noisy Labels (LNL)~\cite{nishi2021augmentation,li2022selective} focuses on designing robust algorithms that are immune to label noise. However, these techniques are less explored in the context of object detection. This work focuses on the problem of bounding box errors in object detection datasets. 
How does the bounding box error term happen? We attribute this phenomenon to annotator fatigue, which means that the occurrence of bounding box errors is directly linked to the individual annotator and not to the individual image. Thus, the error values of the bounding box labels depend on the person who annotate them, and we have no need to care about every bouding box error. There exists an average error value that can reflects the fatigue level of the annotators. Additionally, by investigating the distribution of bounding boxes in the dataset from both global and class-specific perspectives, we find that objects in the dataset are not uniformly distributed: small objects constitute the majority, while large objects are relatively scarce. Therefore, based on the distribution of the dataset, we can categorize objects in the images (e.g., by bounding box size) and then estimate a separate average error value for each category to achieve a more precise bounding box label calibration.

Inspired by these observations, this paper proposes a novel smooth bounding box (SBB) strategy to mitigate the degradation in object detection accuracy resulting from dataset errors, specifically bounding box annotation inaccuracies. The proposed strategy includes two distinct implementation approaches.
On the one hand, we propose to implement our SBB strategy through a global scaling correction method, named GSC, which reformulates the challenging problem of dataset error correction into a more tractable task of estimating an average error value. Firstly, the width and height values of the objects’ bounding box labels are
expanded or contracted by a constant value of $\alpha$ pixels. Secondly, we compute the maximum Intersection over Unio (IoU) among three values: the IoU between the predicted bounding box and the original label, the expanded label, and the contracted label. In this way, GSC decreases the reliance on manual labels during the training process, thereby reducing the occurrence of large loss values caused by label errors. Figure~\ref{fig_smooth_lteration} demonstrates how our method increases the IoU value during anchor box regression by introducing uncertainty to the bounding box labels.
On the other hand, we propose to implement our SBB strategy through an interval scaling correction method, named ISC, which assigns a smooth factor to objects whose bounding boxes are located within different intervals.
The smoothing operation for each object is the same with method GSC. To solve the problem of finding the optimal smooth factor combination in high dimensional space, we design a model fine-tuning strategy. 
To sum up, the major contributions of this study are articulated in the following points:
\begin{itemize}
\item We propose a SBB strategy to mitigate the decline in detection accuracy caused by dataset errors. 
To this end, we popose a global scaling correction method (i.e., GSC), to evaluate the average value of bounding box annotation errors, and transform it into a optimal parameter search problem. Our method is directly applied to the bounding box loss, enabling higher detection accuracy for the model without the need for an additional head or modules that would compromise detection speed.
	
\item We propose a SBB method based on interval scaling (i.e., ISC), which partitions objects into non-overlapping intervals according to their relative sizes and computes an average error value for bounding boxes within each interval individually. Additionally, we design a model fine-tuning strategy that significantly reduces parameter search cost. As a result, ISC achieves more precise and accurate detection results compared to global error estimation methods.

\item Our SBB strategy improves the model's robustness to bounding box errors and stabilizes the loss values during training. To quantitatively evaluate the degree of smoothness, we introduce a metric called Normalized Residual Fluctuation (NRF), which is based on a sliding window residual calculation mechanism.

\item We evaluate our proposed method on three established baseline models. Experimental results demonstrate that across various system configurations, both our proposed components, GSC and ISC, consistently achieve significant improvements in detection accuracy over current state of the art methods, including YOLOv11n.
\end{itemize}

The remainder of this paper is structured as follows. Section~\ref{RelatedW} reviews mainstream existing methods in object detection. Section~\ref{method} provides a detailed description of the proposed approach, including both GSC and ISC. Experimental results and corresponding analyses are presented in Section~\ref{Experi}. Finally, Section~\ref{conclu} concludes the paper by summarizing our contributions and suggesting potential directions for future research.

\section{RELATED WORK}
\label{RelatedW}

With the advances of CNNs, object detection algorithms have made a breakthrough. According to the implementation process, object detection can be mainly divided into two categories: two-stage methods and one-stage methods. 

Two-stage methods, known for their high accuracy, trace their origins to R-CNN \cite{girshick2015region}. Subsequently, SPP-Net \cite{he2015spatial} and Fast R-CNN \cite{girshick2015fast} improved efficiency by proposing RoI Pooling. Thereafter, Mask R-CNN \cite{he2017mask} extended Faster R-CNN with an additional instance segmentation branch and proposed RoI Align. Meanwhile, FPN \cite{lin2017feature} enhanced small-object performance through multi-scale feature fusion.Numerous variants have followed, such as Libra R-CNN \cite{pang2019libra} for addressing sample imbalance, DetectoRS \cite{qiao2021detectors} for strengthening feature representation, PETDet \cite{li2023petdet} for optimizing fine-grained detection, and AFPN \cite{yang2023afpn} for improving the efficiency of non-adjacent layer feature fusion. Overall, two-stage detection methods, owing to their accuracy advantages, have long maintained leading performance on benchmark datasets such as VOC and COCO, constituting one of the mainstream technical routes in the field of object detection.

In parallel, one-stage detectors pursued simplicity and speed by directly predicting objects in a single pass. Starting with You Only Look Once (YOLO) \cite{redmon2016you}. Subsequent variants addressed different issues, such as the introduction of anchor boxes and multi-scale training in YOLOv2 \cite{redmon2017yolo9000} and SSD \cite{liu2016ssd}, the use of FPN to improve small-object detection in YOLOv3 \cite{redmon2018yolov3}, systematic engineering optimization and community-driven design in YOLOv4 \cite{bochkovskiy2020yolov4}, and loss rebalancing in RetinaNet \cite{lin2017focal}. Later versions further enhanced accuracy and lightweight design \cite{wang2023yolov7,wang2024yolov9}, while YOLOv10 \cite{wang2024yolov10}, eliminating the need for non-maximum suppression (NMS), marked the beginning of a fully end-to-end YOLO framework. TOOD \cite{feng2021tood} further refined the scoring mechanism by integrating class confidence with regression quality.

Beyond architectural differences, both paradigms have benefited from improvements in localization accuracy and prediction head design. A series of IoU-based regression losses—GIoU \cite{rezatofighi2019generalized}, DIoU \cite{zheng2020distance}, CIoU \cite{zheng2021enhancing}, and EIoU \cite{zhang2022focal}.Meanwhile, auxiliary prediction heads, such as IoU-aware heads \cite{jiang2018acquisition}, centerness head \cite{tian2019fcos}, density estimation head \cite{Chu_2020_CVPR} and explicitly modeling object relations within the head \cite{zhuang2018relationnet}.

While architectural and regression improvements have significantly advanced detection performance, the reliability of training labels has emerged as a critical bottleneck—especially in large-scale or crowdsourced datasets where annotation errors are prevalent.In the denoising area, where annotation errors and label noise have become increasingly critical issues, various solutions have emerged. Crowdsourcing\cite{krishna2016embracing,anjum2021exploring} and multi-stage validation tasks for relabeling bounding boxes \cite{penquitt2025label,hu2019crowdsourcing} are driven by human beings. Cleaning frameworks \cite{bernhardt2022active} prioritize samples for relabeling by comparing the estimated correctness of labels with task difficulty. These include CROWDLAB \cite{goh2022crowdlab}, CrowdBA \cite{guo2025crowdba} and Crowd R-CNN \cite{hu2019crowd}, or a more direct approach is the release of new “cleaned” datasets like COCO \cite{lin2014microsoft} and PASCAL+ \cite{kim2024active}.

To reduce dependence on manual correction, recent approaches aim to enhance model robustness through data augmentation or self-supervision. One promising direction injects synthetic noise during training—such as bounding-box localization noise \cite{liu2022towards}, Gaussian perturbations applied to object positions \cite{bernhard2021correcting}, and geometric transformations of bounding boxes (e.g., scaling, rotation, and translation). Additionally, cross-domain transfer learning has been investigated in noisy existing environments \cite{liao2024transferring}.

Another line of work adapts label smoothing—originally developed for classification \cite{muller2019does, szegedy2016rethinking} to the detection setting. For example, teacher-student frameworks use softened pseudo-labels from a teacher network to guide student training \cite{xu2021end,tang2021humble},softening the contribution of RoIs \cite{wu2020soft}.A few recent approaches attempt to simulate soft labels from single-class annotations using deep learning techniques \cite{li2024probability, gao2024ps, nguyen2022improving}.

\section{PROPOSED METHOD}
\label{method}
\subsection{Dataset Error}
\label{Dataset Error}
In object detection datasets, annotation errors frequently occur due to the inherent challenges associated with manual labeling. Factors such as variations in object sizes, occlusion, and complex backgrounds complicate the task of accurately drawing bounding boxes. Small or overlapping objects are especially susceptible to inaccurate localization. Furthermore, as human annotators may experience fatigue or develop subjective biases over time, this can result in inconsistent or noisy labels. 
Therefore, this paper focuses on mitigating the decline in detection accuracy caused by annotation errors in the dataset.

\begin{figure}[t]
\centering
\includegraphics[width=0.48\textwidth]{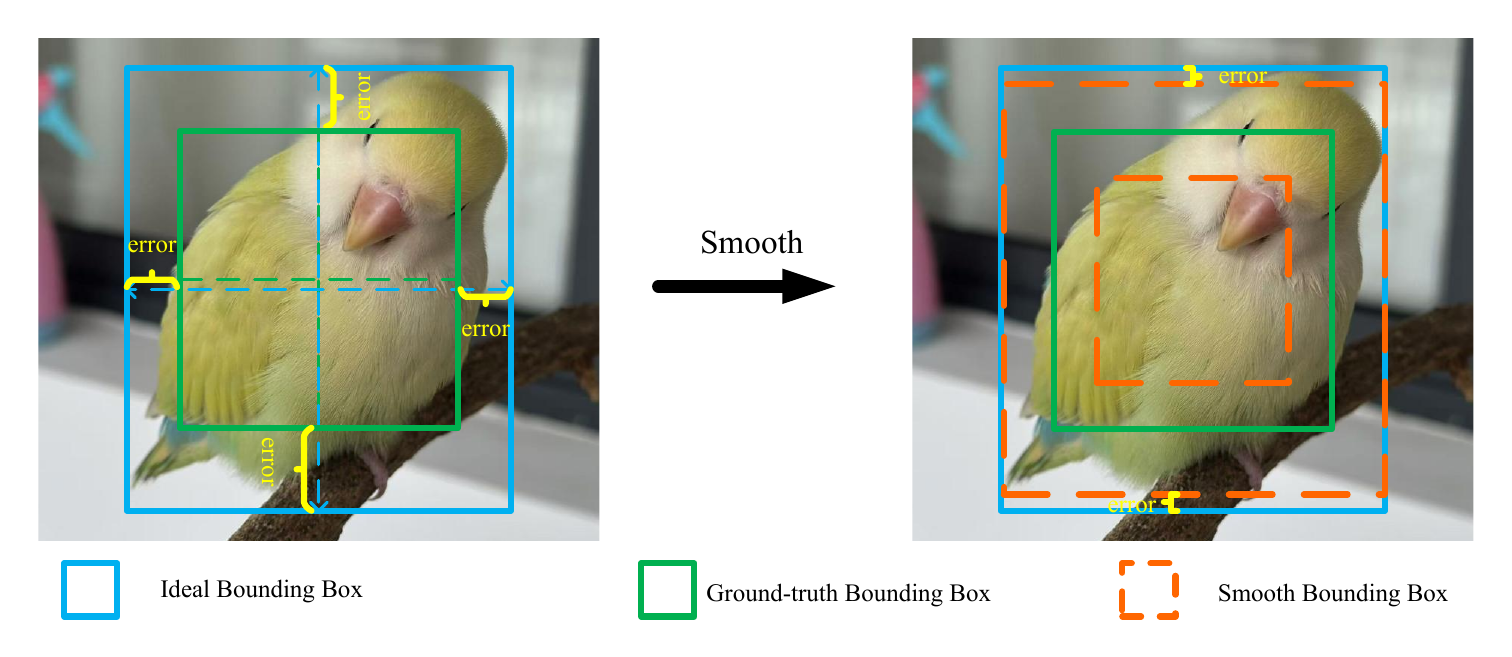}
\caption{Schematic of the dataset error and the proposed SBB strategy. The annotated ground truth is smoothed in two ways: expansion and contraction (indicated by the orange bounding boxes), which brings the expanded bounding box closer to the ideal bounding box, effectively minimizing annotation errors.
}
\label{fig_error}
\vspace{-0.2cm}	
\end{figure}

Defind the dataset as $\mathcal{D}_{\text{raw}}$:
\begin{align}
 \begin{split}
\mathcal{D}_{\text{raw}} &= \{(\mathbf{Img}_i, (C_{i1}, \bm{b}_{i1}), (C_{i2}, \bm{b}_{i2}),..., (C_{im},\bm{b}_{im}))\}_ {i=1}^n,\\ 
& \text{s.t.} \,\,\bm{b}_{ij} = (x,y,w,h),
 \,\,j=1,...,m, \\
 \end{split}
\end{align}
where $\mathbf{Img}_i$ is the $i^{\text{th}}$ input image; $m$ is the number of objects in the $\mathbf{Img}_i$ and $C_{ij}$ is the classification category of the $j^{\text{th}}$ object; $\bm{b}_{ij}$ is the $j^{\text{th}}$ bounding box label (annotated ground truth, the green bounding box in Figure \ref{fig_error}) with an error in the $\mathbf{Img}_i$, which includes the coordinates $(x, y)$ of the upper-left corner, as well as the width $w$ and height $h$ of the bounding box, respectively;  $n$ is the number of images in the dataset.
Since our research focuses on the bounding box, we igore the class labels. Actually, identifying errors for all the bounding box labels is impractical. How does the error term happen? We attribute this phenomenon to annotator fatigue, suggesting that the occurrence of bounding box errors is directly linked to the individual annotator and not to the individual image. Based on this, we have no need to care about each bounding box error, and there exists an average error value that can reflect the fatigue level of the annotators. We can use this average value to decrease most of the errors, and improve the overall quality of the supervision signal.

Denoting the ideal bounding box label as $\tilde{\bm{b}}_{ij}$ (i.e., a label with completely correct annotations and no errors), we calculate $\varepsilon_{ij}$ with $\varepsilon_{ij} \leftarrow (\tilde{\bm{b}}_{ij} - \bm{b}_{ij})$, an expression that represents the average of the sum of the bounding box's errors in all four directions (top, bottom, left, and right), as shown in the left subfigure of Figure~\ref{fig_error}. 
Therefore, the error of the bounding box label for $\mathbf{Img}_i$, denoted as $\varepsilon_{i}$, can be calculated by Formula~\eqref{eq:error}:
\begin{align}
\label{eq:error}
\varepsilon_{i} &= \sum^{m}_{j=1}
\varepsilon_{ij}.
\end{align}
That means summing the errors of all the $m$ bounding boxes. 
The average error value of the dataset, denoted as $\varepsilon_{v}$, is obtained by averaging across all images, as follows:
\begin{align}
\label{eq:EV}
\varepsilon_{v} &=\frac{1}{n}\frac{1}{m} \sum^{n}_{i=1}\varepsilon_{i}.
\end{align}
Thus, the objective of our SBB strategy is to identify an average error value $\varepsilon_{v}$ that shifts the annotated bounding boxes closer to their corresponding ideal labels $\bm{\tilde{b}_{ij}}$.

\subsection{Smooth Bounding Box}
\label{smootbb}
In actual model training task, hard labels, i.e., the one-hot vector, can cause the model to overfit the training set. According to the definition of smooth labels~\cite{muller2019does, szegedy2016rethinking} in classification task, the label smoothing method modifies the hard label by distributing a small portion of the confidence uniformly across all classes. The process can be described as follows:
\begin{align}
y_i^{LS} &= y_i(1-\alpha) + \frac{\alpha}{K}, \,\,i=1,...,K,
\end{align}
where $y_i$ and $y_i^{LS}$ are the $i^{\text{th}}$ true class label and the corresponding smooth label, respectively; $\alpha$ is a smooth factor that usually set to 0.1; $K$ is the number of classification categories. This formula can also be expressed as follows:
\begin{align}
y_i^{LS} & =\left\{
                \begin{array}{ll}
                  1-\alpha, \,\,\, \textrm{if} \,\,i== \textrm{true class},\\
                  \frac{\alpha}{K}, \,\,\, \textrm{others}.
                \end{array}
              \right.
\end{align}
The key idea behind label smoothing is to reduce the certainty of hard labels by introducing a smoothing factor. This is done by assigning a small amount of probability to other classes, effectively turning absolute “yes” or “no” into softer ``probably yes” or ``probably no.” For instance, if $\alpha=0.1$, the true class label will be adjusted to 0.9, while the other classes are smoothed to 
$\frac{0.1}{K}$.

Since the bounding box labels contain annotation errors as described in Section \ref{Dataset Error}, blindly trusting these error-containing labels will cause the model training to deviate from the true data distribution. Therefore, to mitigate the impact of bounding box annotation errors ($\varepsilon_{v}$), we propose implementing label smoothing on the bounding box labels.
Specifically, we propose to transform the dataset through a mapping function $f(\bm{b}_{ij},\alpha): \bm{b}_{ij} \rightarrow \bm{b}_{ij}^{s}$, which makes $\varepsilon_{ij} = \bm{b}_{ij}^{s} - \tilde{\bm{b}}_{ij}$ approach 0, thereby ensuring $\varepsilon_{v}$ also approaches 0. Here, $\bm{b}_{ij}^{s}$ denotes the smoothed bounding box label, where $\alpha \in (0,1)$ controls the degree of smoothing. Unlike classification labels that permit unidirectional smoothing only, bounding box labels allow for bidirectional smoothing, meaning expansion and contraction transformations can be applied simultaneously.
Therefore, the bounding box smooth operation defines a region, denoted as $\mathcal{N}_{\alpha}(\bm{b}_{ij})$:
\begin{align}
\mathcal{N}_{\alpha}(\bm{b}_{ij}) \in [(\bm{b}_{ij}-\alpha),(\bm{b}_{ij}+\alpha)],
\end{align}
where $\pm \alpha$ indicates that the bounding box $\bm{b}_{ij}$ is expanded or contracted by $\alpha$ pixels.
If the predicted bounding box values $\bm{b}_{ij}^{pre} \in \mathcal{N}_{\alpha}(\bm{b}_{ij})$, it can be considered that the detection is correct, i.e $\bm{b}_{ij}^{pre}$ == $\tilde{\bm{b}}_{ij}$. 
So, the actual function of the mapping function $f(\bm{b}_{ij},\alpha)$ is to obtain suitable a smooth region. 

With regard to the issue of the average error $\varepsilon_{v}$, we can use the SBB region $\mathcal{N}_{\alpha}(\cdot)$, where $\cdot$ denotes the bounding box label, to encompass it.
Therefore, the error minimization problem is transformed into a region rule design and parameter $\alpha$ search problem.

\subsection{Global Scaling Correction}
\label{smooth constant}
Based on the proposed SBB strategy, we first investigate the GSC scheme. Specifically, we defined the mapping function as:
\begin{align}
\label{equf}
f(\cdot,\alpha) & =\left\{
                \begin{array}{ll}
                  (x,y,w-\alpha,h-\alpha), \,\,\, \textrm{contraction operation},\\
                  (x,y,w+\alpha,h+\alpha), \,\,\, \textrm{expansion operation}.
                \end{array}
              \right.
\end{align}
As shown in Equation \ref{equf}, the width and height values of the objects' bounding box labels are expanded or contracted by a constant value of $\alpha$ pixels. Therefore, we need to adjust the loss function of object detection models, based on the smoothed bounding boxes, where the loss for the bounding box (denote as $L_{box}$) is defined by IoU. The bounding box loss is formulated as follows:
\begin{align}
L_{box} &= 1-IoU.
\end{align}
The IoU value is calculate by:
\begin{align}
IoU(\bm{a},\bm{b})=\frac{|\bm{a}\cap \bm{b}|}{|\bm{a} \cup \bm{b}|},
\end{align}
where $\bm{a}$ represents the predicted bounding box and $\bm{b}$ represents the labeled bounding box, respectively. Since the bounding box label is smoothed in two directions, the calculation of IoU should be adjusted to account for this feature. The adjusted value is referred to as smooth IoU (denoted as ($IoU_s$)).

According to the analysis in Section \ref{smootbb}, it can be concluded that smooth labels introduce uncertainty by assigning values such as `maybe yes' or `maybe no'. The purpose of this is to decrease the reliance on manual labels during the training process, thereby reducing the occurrence of large loss values caused by label errors. Therefore, we propose calculating the maximum IoU, which involves determining the IoU value between the predicted bounding box and the original label (denoted as ($IoU_0$)), the IoU value between the predicted bounding box and the expanded bounding box label (denoted as ($ IoU_{-\alpha}$)), and the IoU value between the predicted bounding box and the contracted bounding box label (denoted as ($ IoU_{+\alpha}$)). The final smooth IoU value (i.e. $IoU_s$) is then taken as the maximum of these values. Therefore, $IoU_s$ can be formulated as:
\begin{align}
IoU_s &= \textrm{Max}(IoU_0,IoU_{-\alpha},IoU_{+\alpha}).
\end{align}
Since the IoU metric is smoothed using the maximum IoU value, the computation of the bounding box regression loss $L_{box}=1-IoU_s$ becomes more robust to label noise, effectively mitigating the impact of labeling errors that would otherwise lead to excessively large loss values. Figure \ref{fig_error} illustrates the proposed SBB strategy. It can be seen from Figure \ref{fig_error} that by using two smooth operations, expansion and contraction (indicated by orange bounding boxes), the expanded bounding box in this example is brought closer to the ideal bounding box, effectively minimizing the annotation errors.

In summary, our approach processes dataset errors through SBB operations rather than direct data correction. This method essentially functions as an information filter to minimize the influence of annotation noise. Through this approach, we effectively reformulate the challenging dataset error correction problem into a more tractable hyperparameter optimization task, specifically focused on identifying the optimal smoothing parameter $\alpha$.

\subsection{Interval Scaling Correction}
In Section \ref{Dataset Error}, we have defined the single bounding box error $\varepsilon_{ij}$ and the average error $\varepsilon_{v}$ for specific datasets.  Actually, the difference lies in the range: one pertains to a single cell of the dataset, while the other represents the entire dataset. To avoid dealing with a large number of $\varepsilon_{ij}$, we aim to find an average value $\varepsilon_{v}$. However, due to the non-uniform distribution of object sizes in the dataset and the variation in errors introduced when annotating objects of different sizes, it is crucial to apply distinct scaling factors for SBB of varying sizes, within the constraints of computational resources. 

\begin{figure}[ht]
\centering
\includegraphics[width=0.48\textwidth]{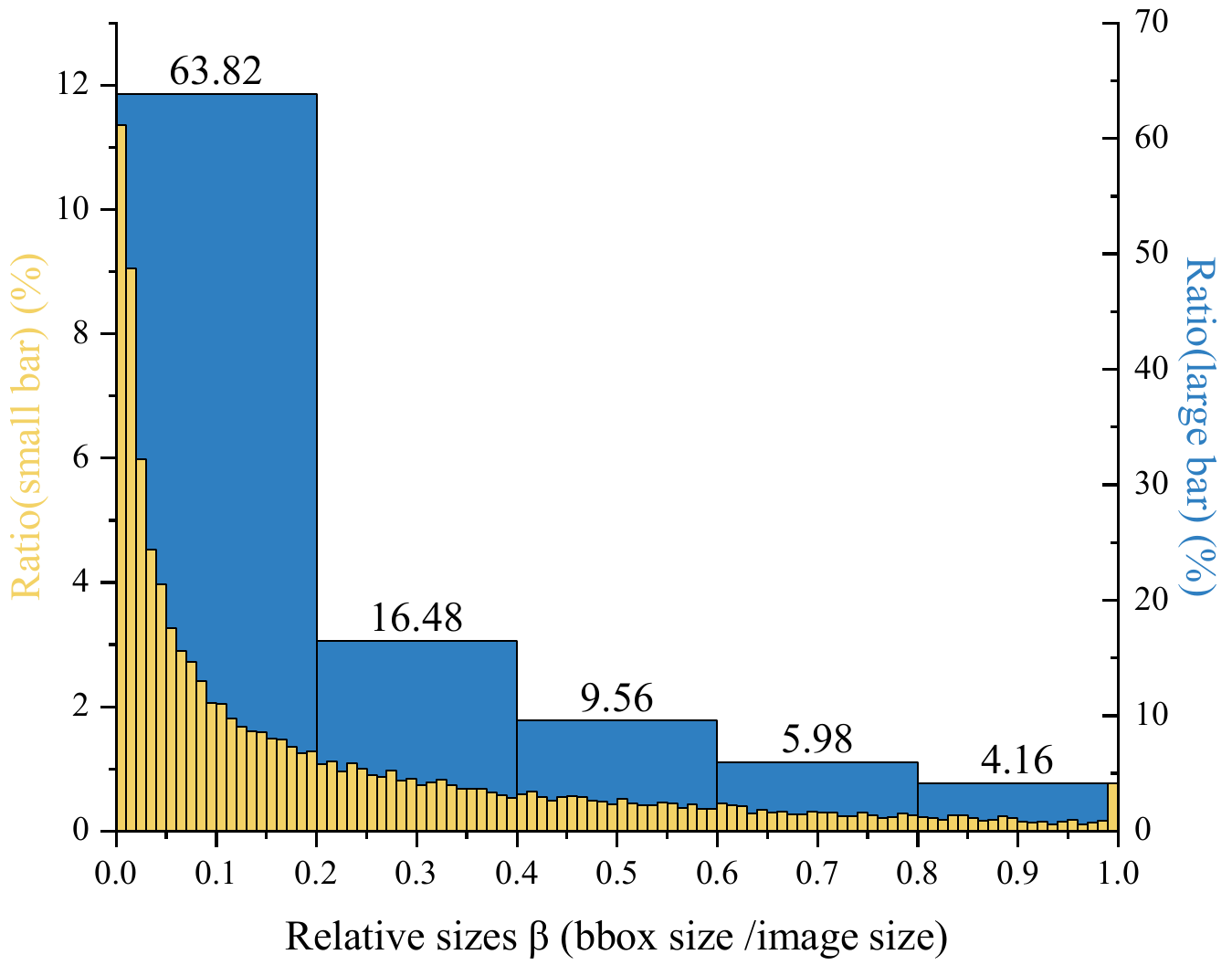}
\caption{Diagram of the object size distribution in the PASCAL VOC dataset, using both coarse and fine interval divisions to represent the dataset distribution. The fine interval divisions split the object into 100 intervals, each with a relative size of 0.01 (represented by the orange bars), while the coarse interval divisions divide the object into 5 intervals, each with a relative size of 0.2 (represented by the blue bars). The horizontal axis represents the relative size $\beta$; the orange bars correspond to the left vertical axis, while the light blue bars correspond to the right vertical axis.}
\label{fig_datasets}
\end{figure}

\begin{figure*}[tp]
    \centering

    \subfloat[Distribution of class bottle]{
        \begin{minipage}[b]{0.31\textwidth}
            \centering
            \includegraphics[width=\linewidth]{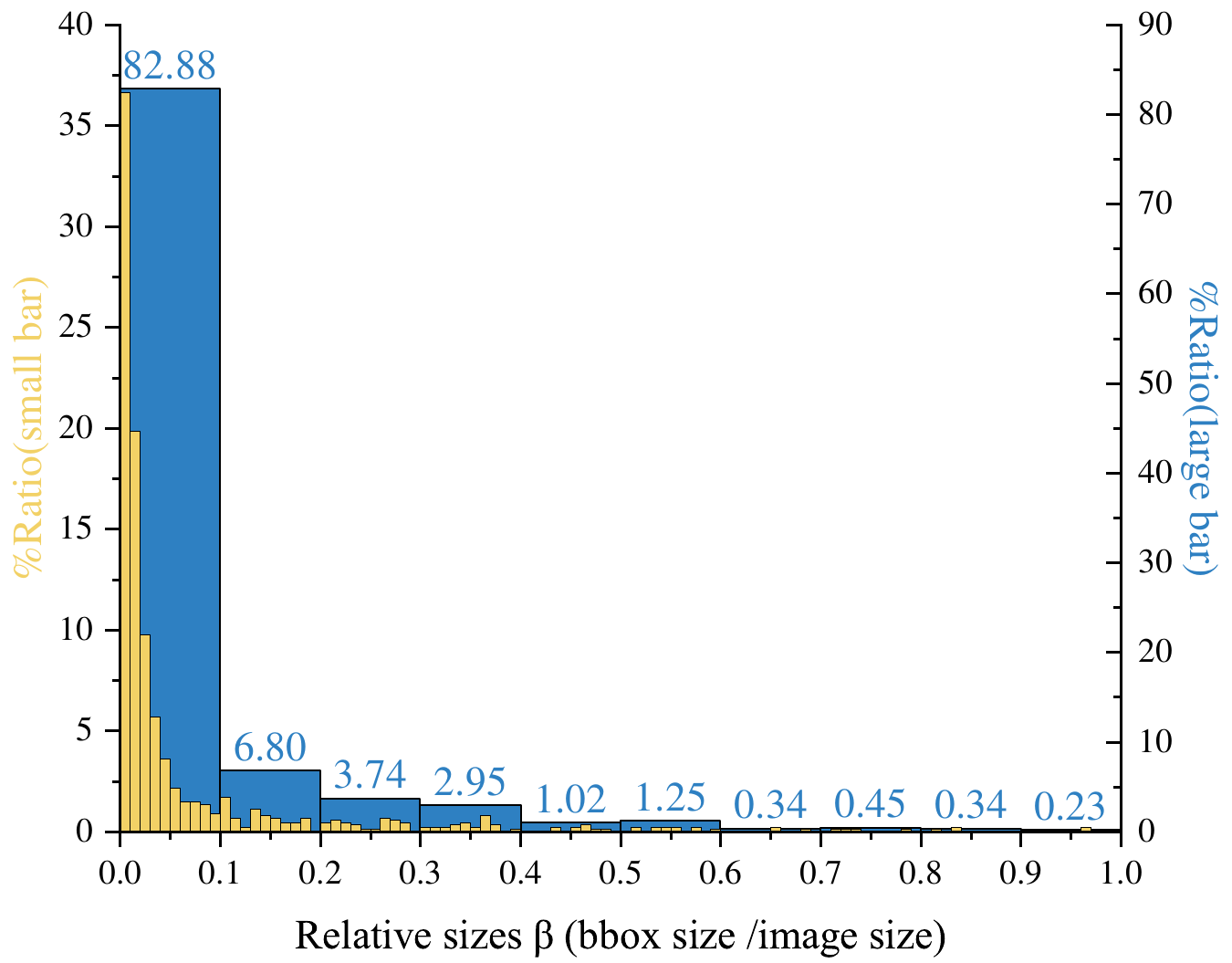}
            \label{fig:constant_scaling1}
        \end{minipage}
    }
    \hfill
    \subfloat[Distribution of class bicycle]{
        \begin{minipage}[b]{0.31\textwidth}
            \centering
            \includegraphics[width=\linewidth]{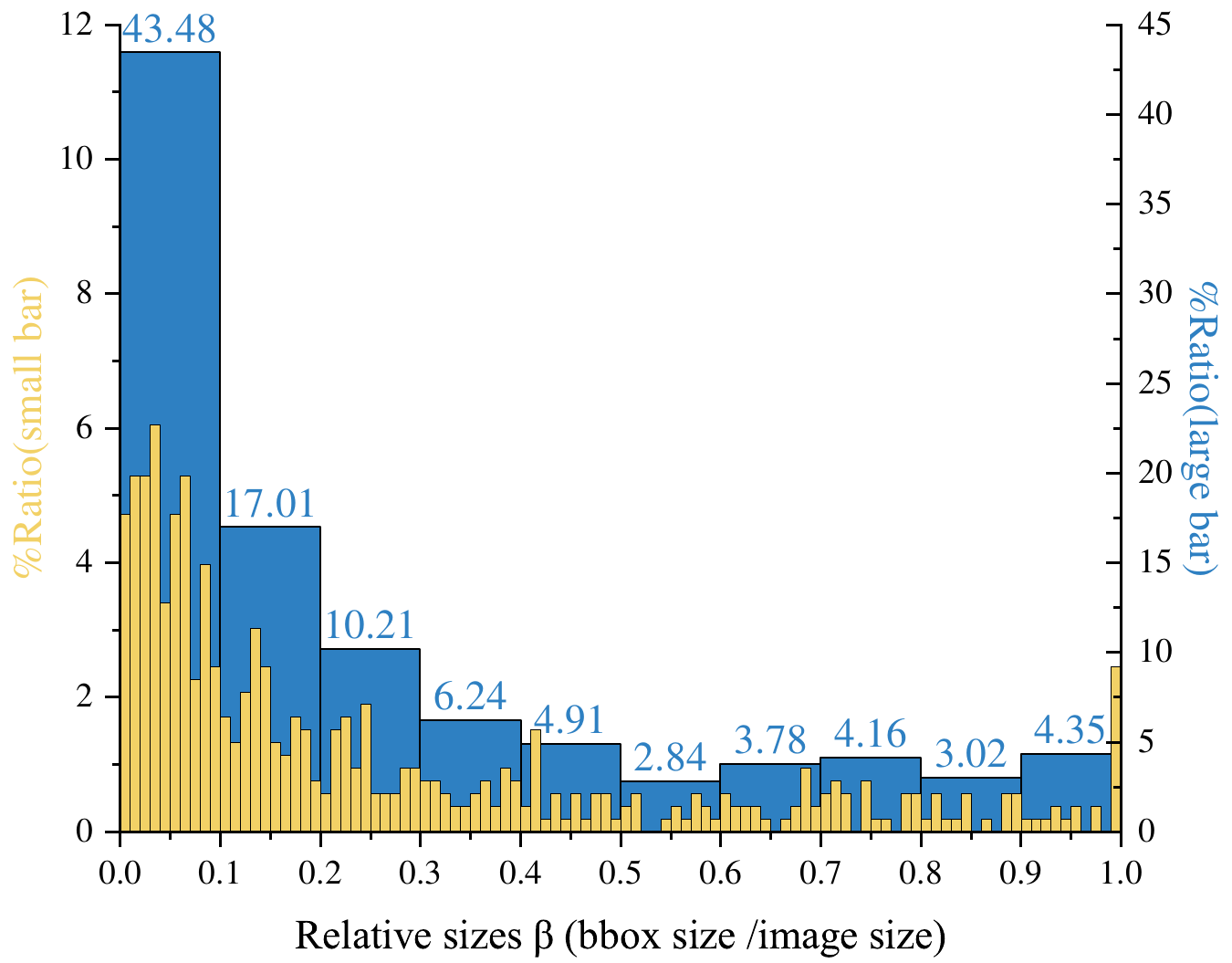}
            \label{fig:constant_scaling2}
        \end{minipage}
    }
    \hfill
    \subfloat[Distribution of class cow]{
        \begin{minipage}[b]{0.31\textwidth}
            \centering
            \includegraphics[width=\linewidth]{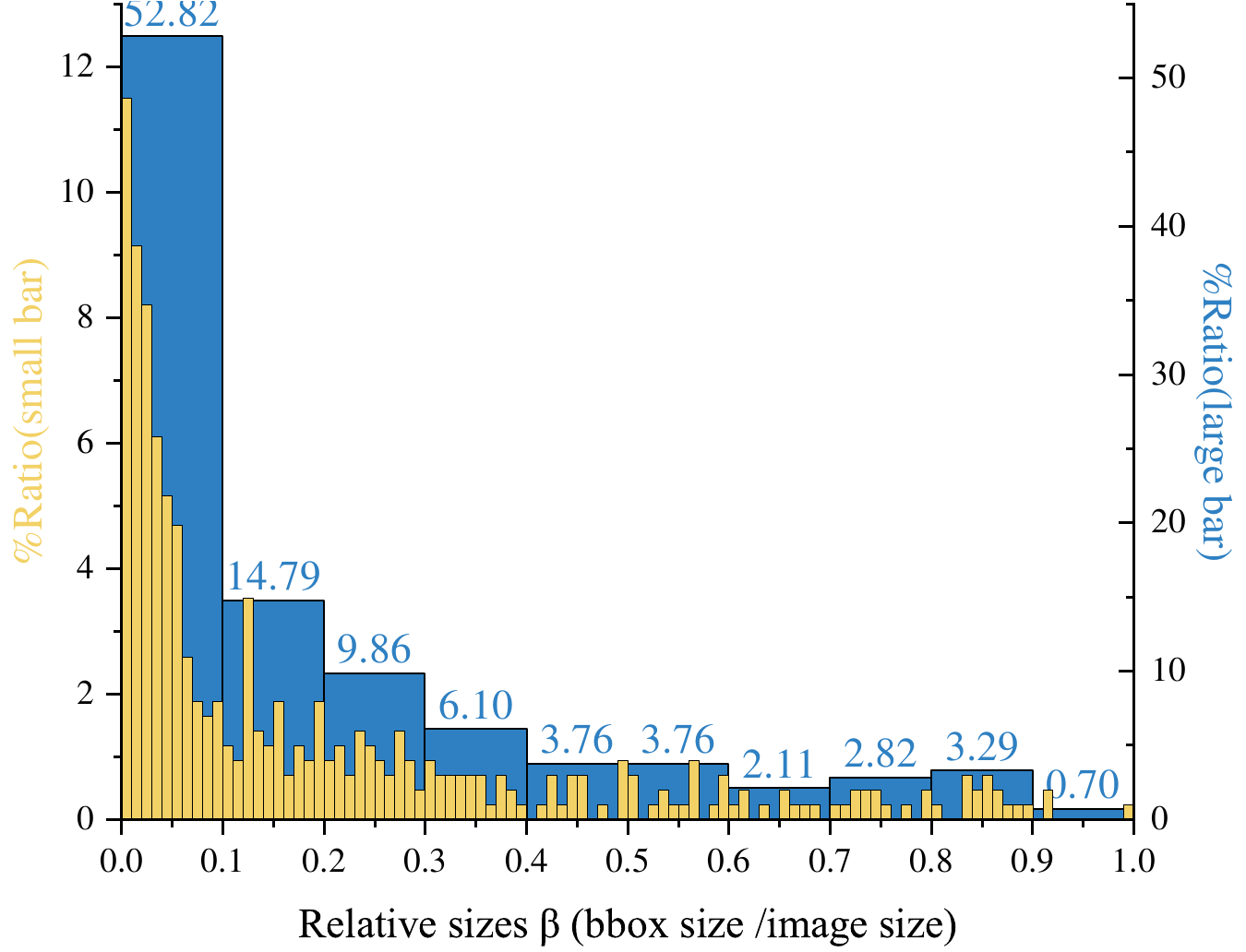}
            \label{fig:constant_scaling3}
        \end{minipage}
    }

    \vspace{0.5em} 

    \subfloat[Distribution of class pottedplant]{
        \begin{minipage}[b]{0.31\textwidth}
            \centering
            \includegraphics[width=\linewidth]{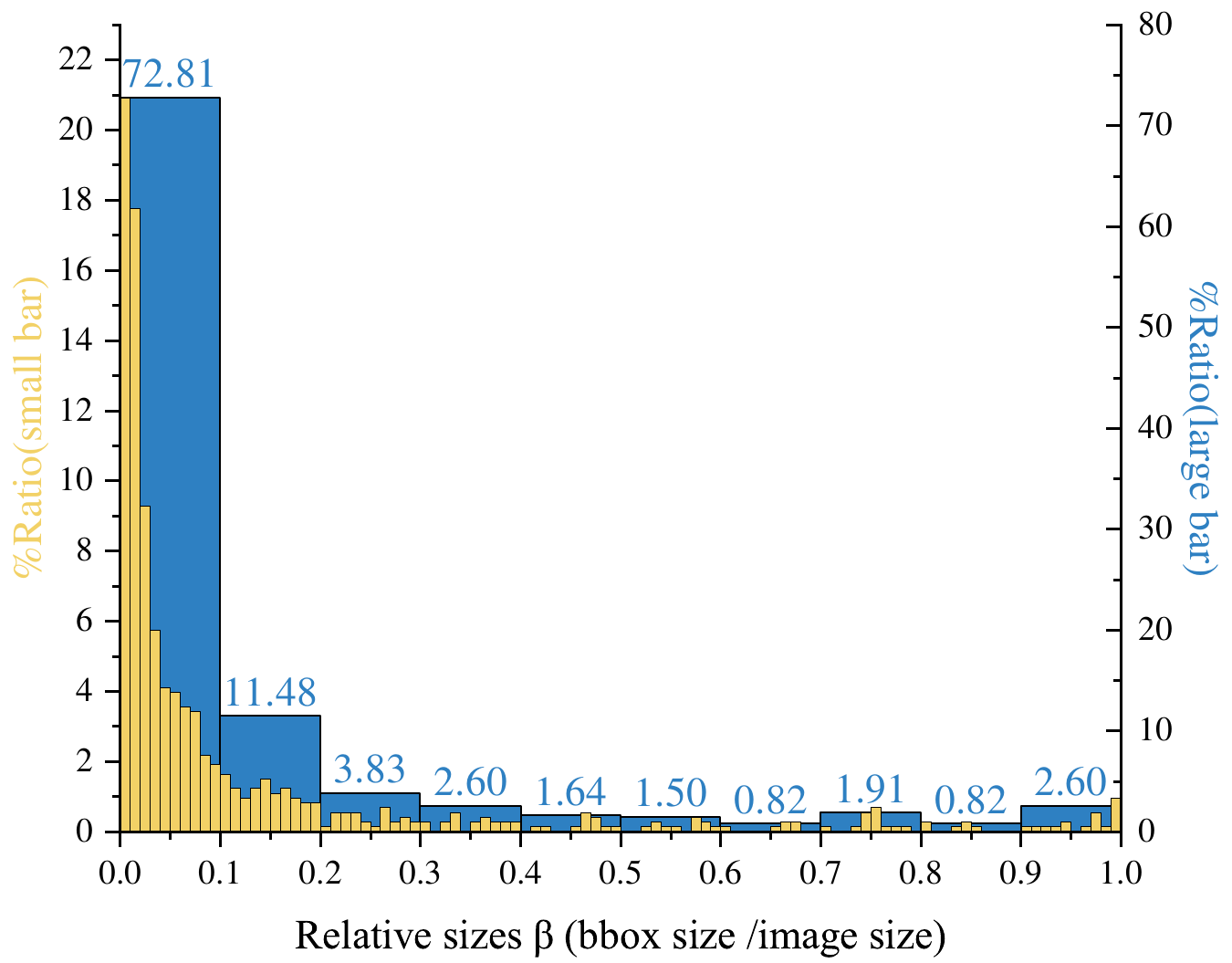}
            \label{fig:constant_scaling4}
        \end{minipage}
    }
    \hfill
    \subfloat[Distribution of class train]{
        \begin{minipage}[b]{0.31\textwidth}
            \centering
            \includegraphics[width=\linewidth]{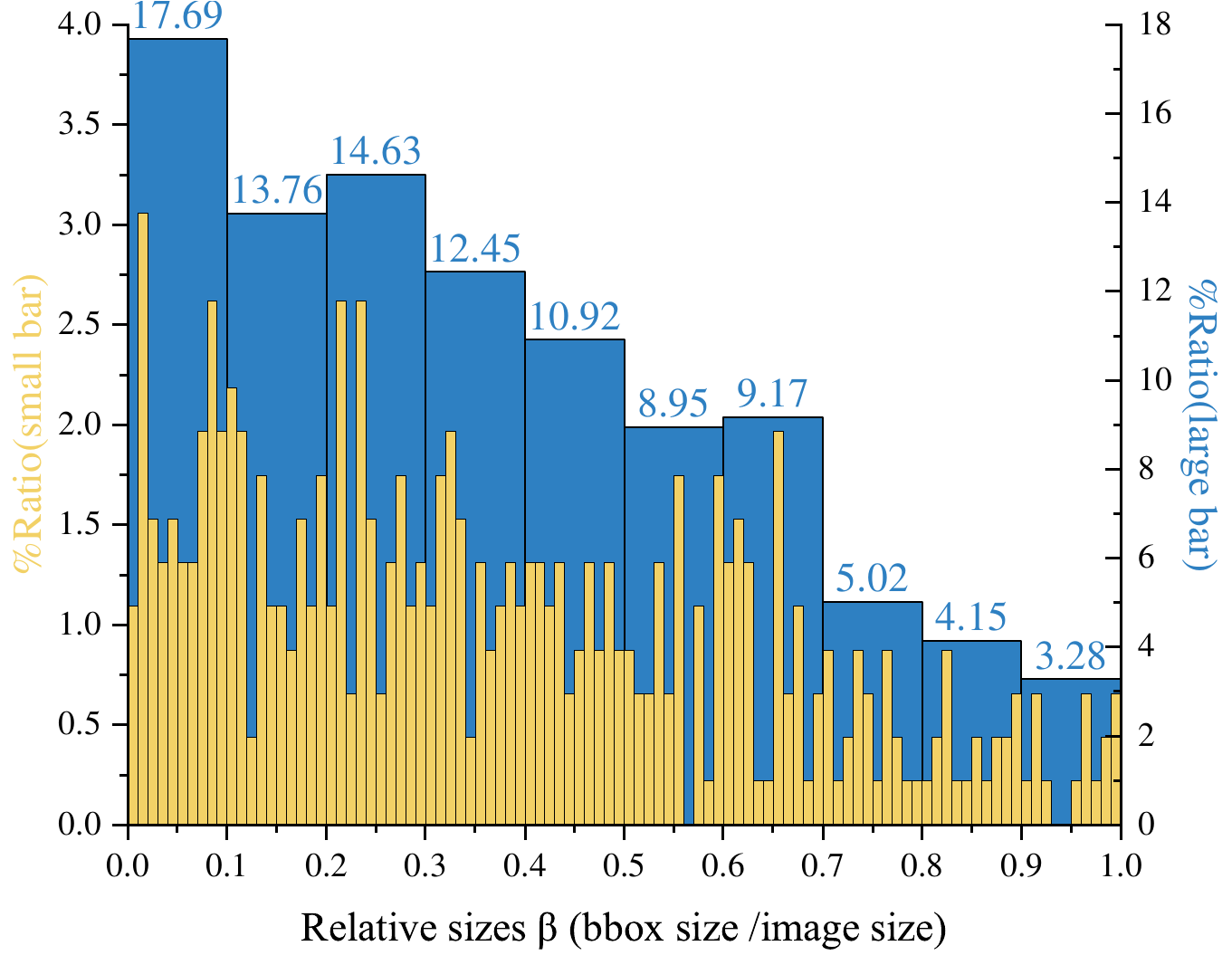}
            \label{fig:constant_scaling5}
        \end{minipage}
    }
    \hfill
    \subfloat[Distribution of class sheep]{
        \begin{minipage}[b]{0.31\textwidth}
            \centering
            \includegraphics[width=\linewidth]{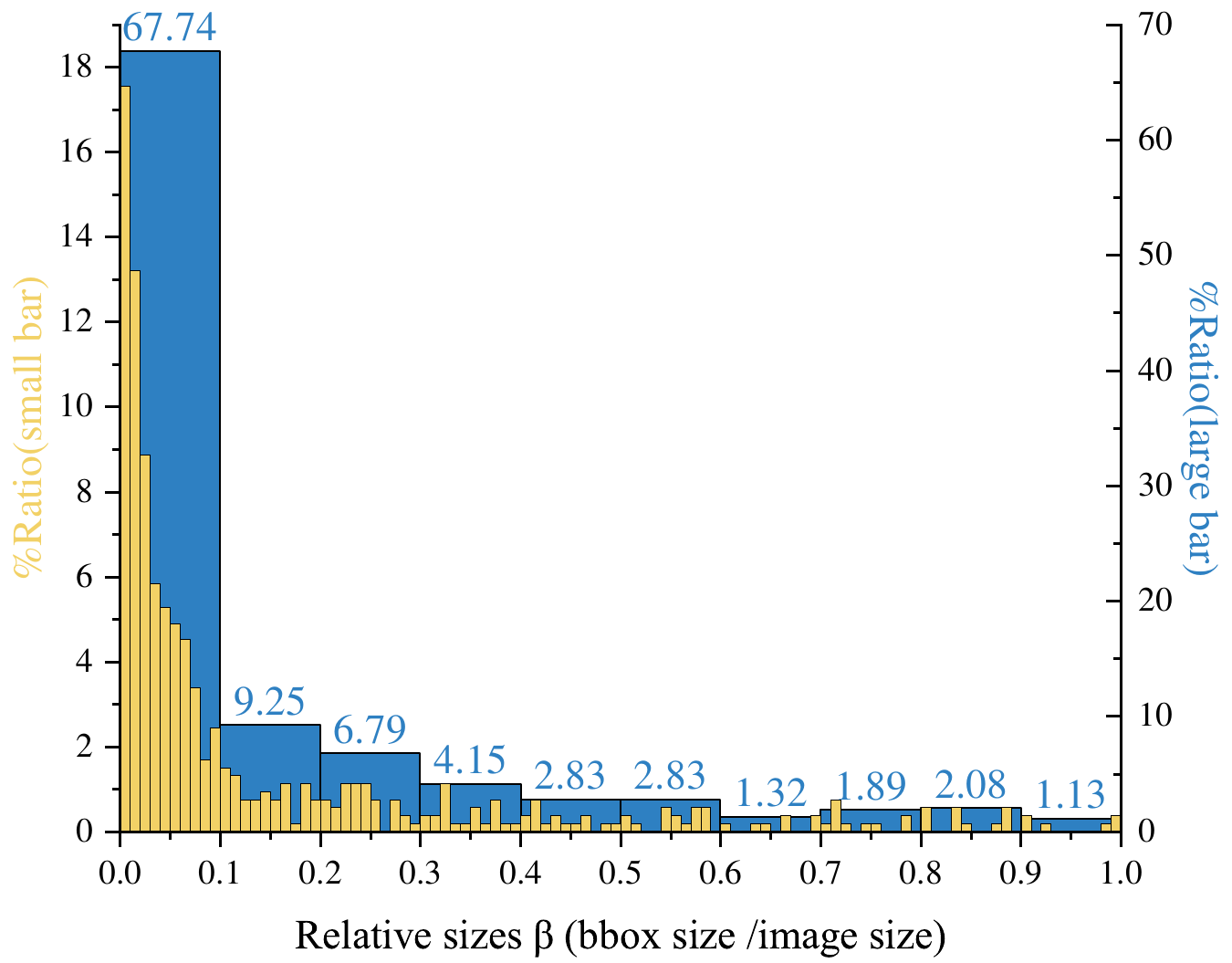}
            \label{fig:constant_scaling6}
        \end{minipage}
    }

    \vspace{0.5em}

    \subfloat[Distribution of class TV]{
        \begin{minipage}[b]{0.31\textwidth}
            \centering
            \includegraphics[width=\linewidth]{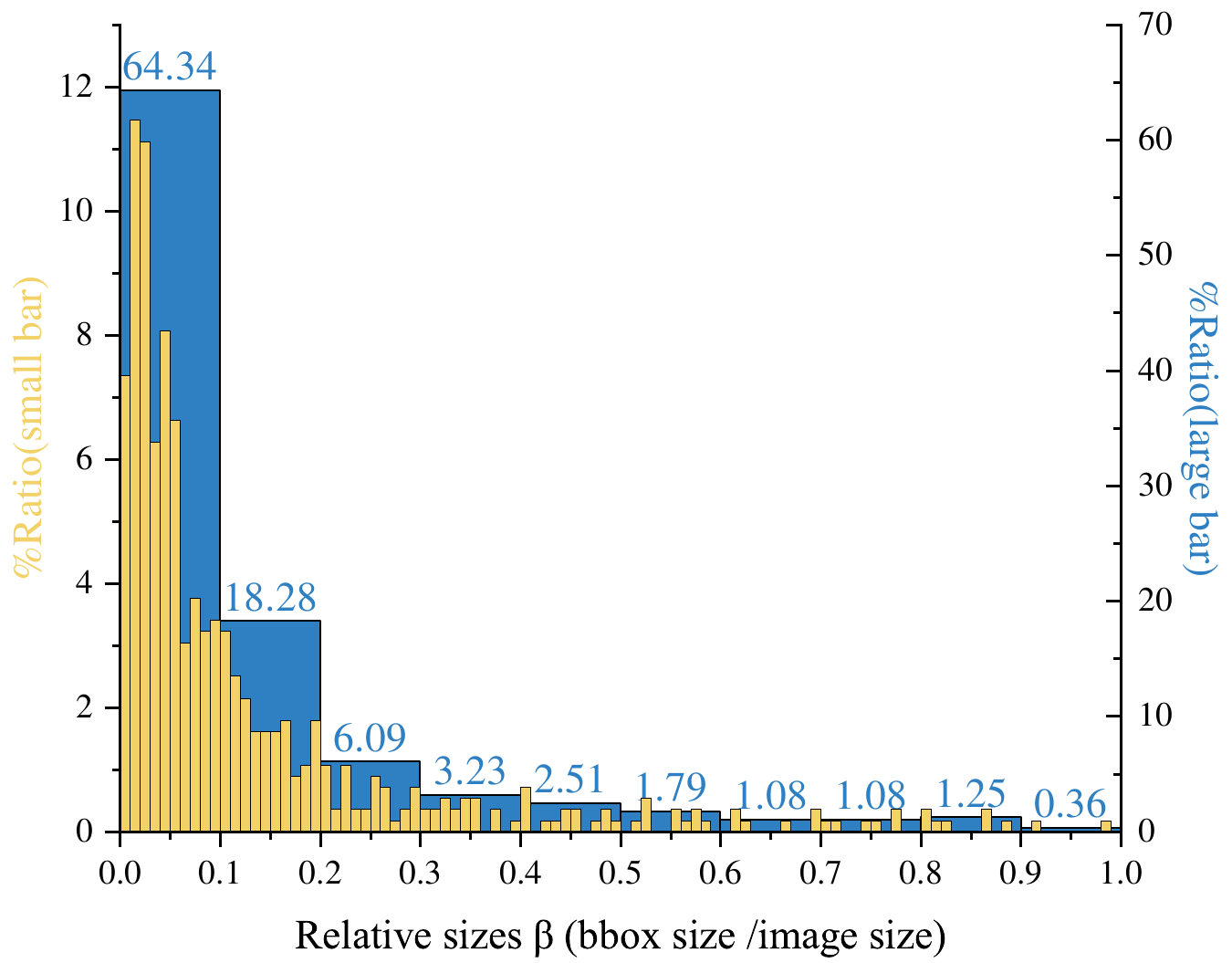}
            \label{fig:constant_scaling7}
        \end{minipage}
    }
    \hfill
    \subfloat[Distribution of class cat]{
        \begin{minipage}[b]{0.31\textwidth}
            \centering
            \includegraphics[width=\linewidth]{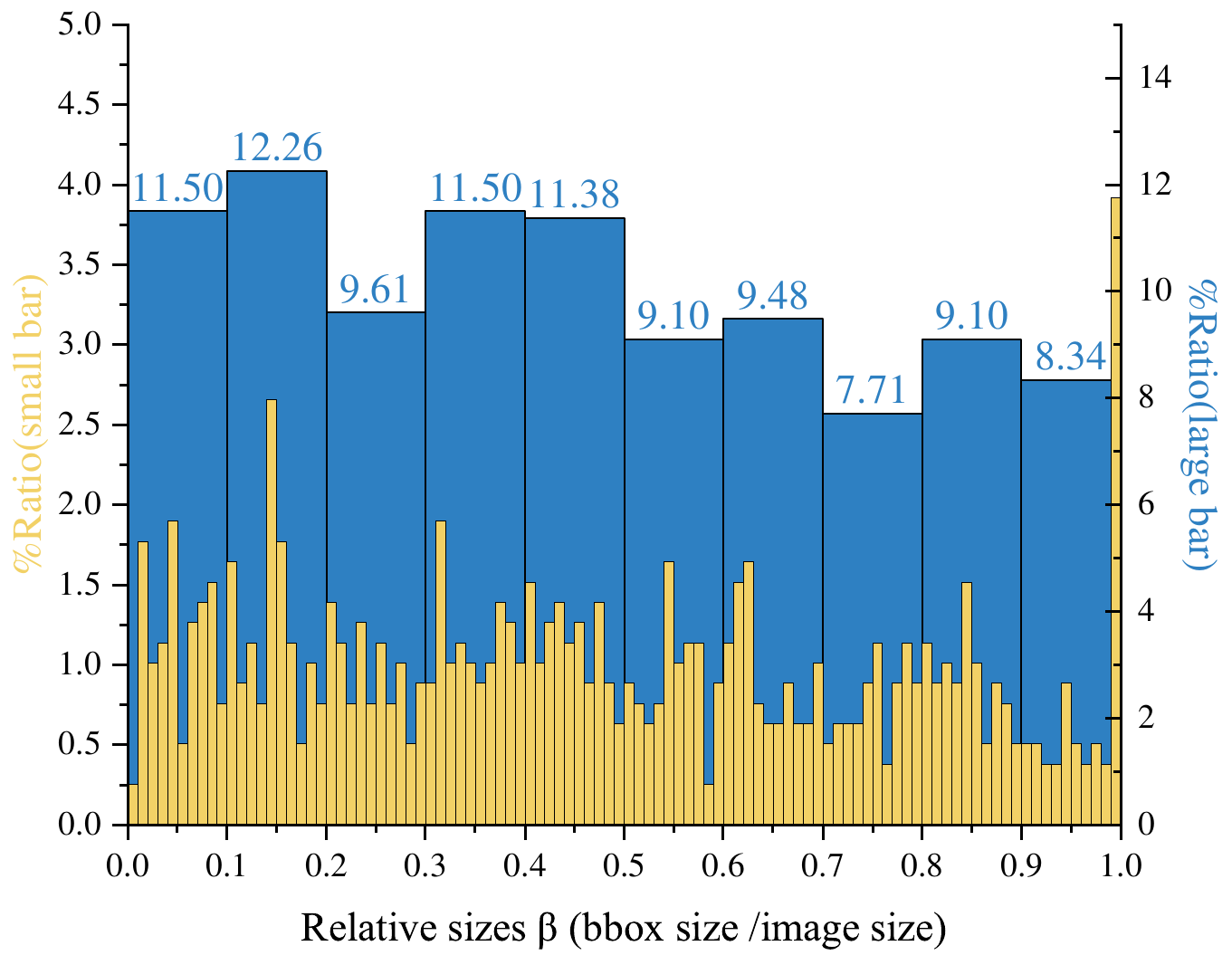}
            \label{fig:constant_scaling8}
        \end{minipage}
    }
    \hfill
    \subfloat[Distribution of class dog]{
        \begin{minipage}[b]{0.31\textwidth}
            \centering
            \includegraphics[width=\linewidth]{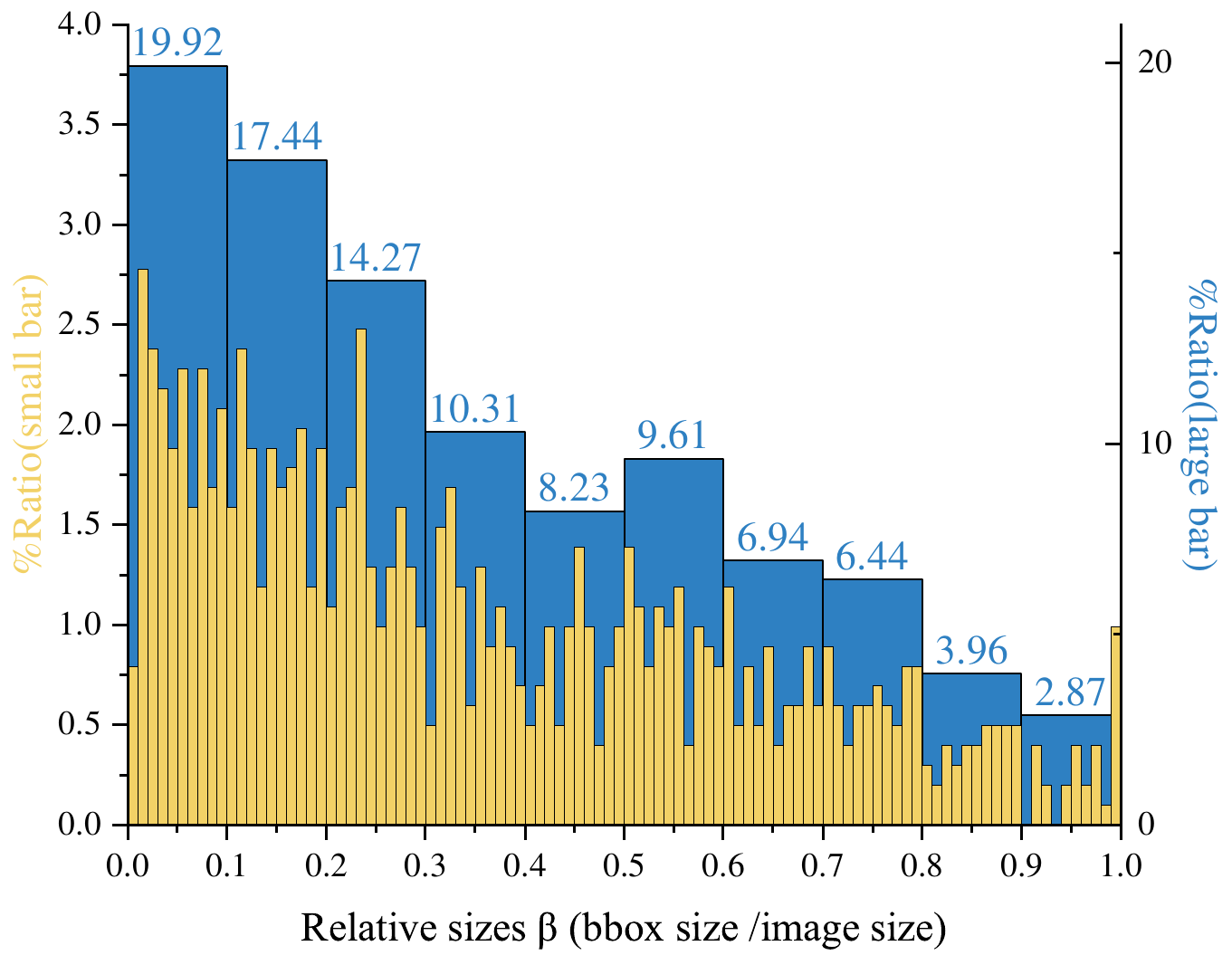}
            \label{fig:constant_scaling9}
        \end{minipage}
        
    }

    \caption{The relative size distribution across multiple classes in the PASCAL VOC dataset, including nine categories namely bottle, bicycle, cow, pottedplant, train, sheep, TV, cat and dog. The fine interval divisions split the object into 100 intervals, each with a relative size of 0.01 (represented by the orange bars), while the coarse interval divisions divide the object into 10 intervals, each with a relative size of 0.1 (represented by the blue bars). The horizontal axis represents the relative size $\beta$; the orange bars correspond to the left vertical axis, while the light blue bars correspond to the right vertical axis. The object relative sizes follow an uneven distribution, with most objects concentrated within the narrow [0, 0.1] range, exhibiting a pronounced unimodal peak.}
    \label{fig:class_distribution}
    \vspace{-0.3cm}	
\end{figure*}

\subsubsection{Dataset Error Distribution} 
There are differences in image and object sizes across different datasets, or even within the same dataset. 
Considering that IoU is a ratio that calculates relative size, and that varying bounding box sizes cannot be standardized across images, we define this problem using the proportion of bounding box size to image size to represent relative sizes, which can be denoted as $\beta$ and formulated as follows:
\begin{align}
\beta = \frac{S_{\text{bbox}}}{S_{\text{img}}},
\end{align}
where $\beta\in [0, 1]$, $S_{\text{bbox}}$ is the size of the object and $S_{\text{img}}$ is the size of image. From the SBB experiments, 
we discover that objects of different sizes exhibit varying effects when the parameter $\alpha$ is changed, indicating that the distribution of the datasets also influences the distribution of the errors. 
In Figure \ref{fig_datasets}, we apply interval partitioning with two scales (0.01 and 0.2 relative sizes as intervals) to statistically analyze the object size distribution of the PASCAL VOC dataset. As shown, the majority of objects are small-scale, with more than half falling within the range 
$\beta\in[0, 0.2]$. The proportion of large-scale objects is minimal, for example, objects with $\beta\in[0.8, 1.0]$ account for only 4.16\%. Additionally, we analyze the relative size distribution across multiple classes (Figure~\ref{fig:class_distribution}). As an illustrative example, we present statistics for nine categories, including bottle, bicycle, cow, and others. As reflected in the comprehensive dataset distribution, Figure~\ref{fig:class_distribution} clearly illustrates that object relative sizes follow an uneven distribution, with most objects concentrated within the narrow [0, 0.1] range, exhibiting a pronounced unimodal peak. These findings indicate that adopting a size-aware approach, specifically by addressing localization errors within the small-scale interval ([0, 0.1]), may significantly improve the method's performance.

According to the relative size, the entire dataset consists of objects whose bounding box sizes fall within the range of $B \in [0, 1]$. Therefore, we can split the objects to $o$ non-overlapping intervals, denoted as $B=\{B_1,B_2,...,B_o\}$, where
\begin{align}
\quad B_i \cap B_j = \varnothing, \,\, i \ne j\,\, \textrm{and} \,\,\{i,j\} \leq o.
\end{align}
Similarly, the overall dataset average error $\varepsilon_{v}$ defined in Equation~\ref{eq:EV} can be partitioned into $o$ parts:
\begin{align}
\varepsilon_{v}=\{\varepsilon_{v1},\varepsilon_{v2},...,\varepsilon_{vo}\},
\end{align}
where $\varepsilon_{vj},j=1,2,..,o$ denotes the average error of the bounding boxes whose relative sizes fall within the interval $B_j$. In this way, the average error $\varepsilon_{v}$ can be calculated by
\begin{align}
\varepsilon_{v}=\frac{1}{o}\sum^{o}_{j=1}\varepsilon_{vj}.
\end{align}

Inspired by this, we propose determining an average error value, denoted as $\varepsilon_{vj}$, for the bounding boxes in each interval individually, and refer to this approach as SBB via Interval Scaling Correction (ISC). This method effectively brings most of the bounding boxes closer to the ideal label. By estimating bounding box errors separately for different intervals based on the distribution of bounding boxes in the dataset, ISC can provide more precise and accurate results compared to a global error estimation.

\begin{figure*}[t]
    \centering
    \includegraphics[width=1\linewidth]{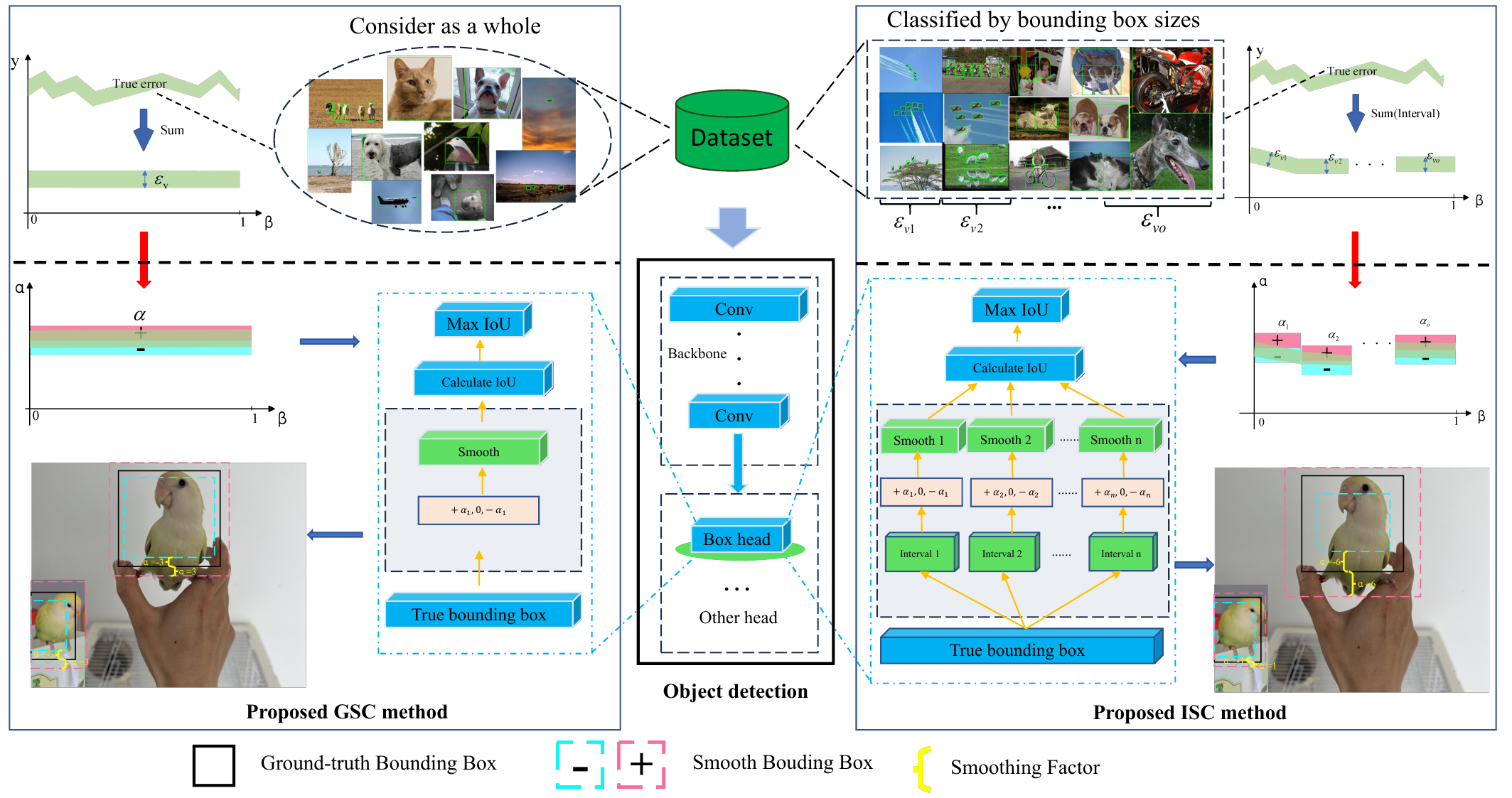}
    \caption{Diagram of the proposed SBB strategy. The left portion shows the global error estimation pipeline for the complete dataset, while the right portion displays the size-stratified local estimation approach. The interval count (hyperparameter $\alpha_o$) can be tuned based on available computing resources during deployment.}
    \label{fig:interval}
\end{figure*}

\subsubsection{SBB via Interval Scaling Correction}
Due to the segmentation of the dataset based on the size of the bounding boxes, we need to restate our SBB theory. 

On the one hand, the mapping function $f(\cdot,\alpha)$ should be redefined as:
\begin{align}
f(\cdot,\alpha_1,\alpha_2,...,\alpha_o),
\end{align}
where $\alpha_1,\alpha_2,...,\alpha_o$ are the smooth factors corresponding to classification objects whose relative sizes fall into the $o^{th}$ predefined intervals.
This implies setting a smooth factor for each interval during model training: Objects with relative size in interval 
$B_i$ are smoothed using the corresponding smooth factor $\alpha_i$, and the smoothing operation can be referred to Formula \ref{equf}. 
On the other hand, the calculation of smooth IoU also needs to be segmented. Using $IoU_s'$ to represent the new smoothed IoU and it can be calculated by follows:
\begin{align}
IoU_s' &= \sum^{o}_{i=1}\textrm{Max}(IoU_0,IoU_{-\alpha_i},IoU_{+\alpha_i}),
\end{align}
Here, the smoothing factor is set to $\alpha_i$ where the relative bounding box size $\beta$ belongs to interval $B_i$. By processing all bounding box labels with their respective smoothing factors to compute individual losses and then summing these values, we obtain the smoothed loss $IoU_s'$.
At this point, the loss for the bounding box can be calculated by $L_{box} = 1-IoU_s'$.

Figure \ref{fig:interval} shows the overall framework of the proposed SBB strategy. In summary, we propose two distinct analytical frameworks: a global error estimation method for the complete dataset, i.e., GSC, and a localized estimation approach stratified by objects' relative size intervals, i.e., ISC,. In practical applications, users can adjust the number of intervals (i.e., the count of hyperparameter $\alpha_1,\alpha_2,...,\alpha_o$) according to available computational resources.

\subsection{Normalized Residual Fluctuation}
\label{nrf}
As explained in Section \ref{smooth constant}, our SBB strategy smooths the IoU value, which directly influences the computation of the bounding box regression loss $L_{box}=1-IoU_s$. As a result, our strategy enhances the model's robustness to bounding box errors, effectively mitigating the impact of labeling inaccuracies that would otherwise lead to excessively large loss values. To capture meaningful information about training stability and learning dynamics without the influence of epoch effects, we extend the classical rolling standard deviation by introducing a task-dependent normalization strategy and propose the Normalized Residual Fluctuation (NRF). 

Let the original sequence be $\bm{L}=\{ l_1,l_2,...,l_N\}$, and divide it into $T$ non-overlapping windows. Assuming the length of each window is $S$, the subsequence of the $t^\text{th}$ window is: 
\begin{align}
\bm{L}_t = \left( l_{(t-1)S+1}, l_{(t-1)S+2}, \dots, l_{tS} \right).
\end{align}
Define the local trend $\hat{\bm{L}_t}^{(S)}$ as the smoothed estimate of $\bm{L_t}$ within a centered window of size $S$. Then, the residual sequence between $\bm{L}_t$ and $\hat{\bm{L}_t}^{(S)}$ can be computed as:
\begin{align}
\bm{r}_t = \bm{L}_t - \hat{\bm{L}_t}^{(S)},
\end{align}
which captures the short-term deviations from the expected learning behavior. 

In practical settings, many training metrics, such as mAP, precision, and recall, naturally fall within the range [0, 1]. However, others, such as loss values, may span an unbounded or arbitrary interval. 
To ensure comparability and scale invariance in our fluctuation measure, we apply optional value-range normalization for metrics whose domain does not lie within [0, 1]. 
Specifically, for a subsequence $\bm{L_t}$ that lies outside this range, it can be normalized as follows:
\begin{align}
\bm{L}_t^{(\text{norm})}[s] = \frac{\bm{L}_t[s] - \min(\bm{L}_t)}{\max(\bm{L}_t) - \min(\bm{L}_t)}, \,\, s = 1,2,...,S.
\end{align}
These residuals exhibit greater fluctuations during unstable learning phases and smaller fluctuations during stable convergence. To quantify this behavior, we calculate the local standard deviation of the residuals equence $\bm{r_t}$, denoted as $\sigma_t$. 
Based on this, the NRF value during the entire model training phase can be expressed as:
\begin{align}
\mathrm{NRF} =
\begin{cases}
\displaystyle \frac{1}{|T|} \sum^{T}_{t=1} 2 \cdot \sigma_t, & \text{if } \bm{L} \in [0, 1], \\
\displaystyle \frac{1}{|T|} \sum^{T}_{t=1} 2 \cdot \sigma_t^{(\text{norm})}, & \text{if } \bm{L} \notin [0, 1], 
\end{cases}
\end{align}
where $\sigma_t$ is computed using the raw residual sequence, and $\sigma_t^{(\text{norm})}$ is computed using the residual equence after value-range normalization.

This approach yields a normalized measure of instability along the model's learning through two complementary normalization strategies. On the one hand, if the metric values do not naturally lie within the [0,1] interval (e.g., loss function values), range normalization is employed to standardize their scales, thereby mitigating biases arising from disparities in magnitude or initial value ranges. On the other hand, 
residual fluctuations around the local trend are quantified using a moving-window standard deviation ($\sigma_t$), providing a scale-invariant measure of training stability across iterations. The envelope defined by $\pm\sigma_t$ (corresponding to approximately 95\% confidence intervals under normality assumptions) characterizes the typical fluctuation range, offering both visual and quantitative insights into the smoothness of the training trajectory.

To sum up, the NRF metric exhibits higher values during periods of rapid, irregular optimization dynamics and lower values as training stabilizes and converges. This property makes NRF particularly valuable for comparing the smoothness of learning trajectories across different training configurations, even when they vary in total epochs or value ranges.

\section{Experimental Results}
\label{Experi}
\subsection{Experimental Setup}
We have evaluated our algorithm on three different object detection models, which are the smallest model of YOLO11 (i.e., YOLO11n\footnote{https://docs.ultralytics.com/zh/models/yolo11/}) \cite{khanam2024yolov11,yolo11_ultralytics}, YOLOv8 \cite{yolov8_ultralytics} and YOLOv5 \cite{yolov5}. In terms of the dataset, we used the classic object detection dataset PASCAL VOC.
Since our approach essentially converts bounding boxes into smooth loss formulations, we trained the models from scratch to properly establish this relationship, rather than relying on fine-tuning, which might preserve undesirable pre-trained features.
To ensure a fair comparison, we maintained identical hyperparameter optimization settings (budget and mechanism) across all baselines and our method. The only exception is the epoch parameter, which we set to 1500 instead of the standard 600 to allow for more thorough training. However, we also allowed extended training for the baselines while using early stopping (with 100-epoch patience) to avoid overfitting and ensure efficiency.
Additionally, due to limitations in experimental equipment, we were unable to conduct tests across different GPU models (e.g., RTX 4090 and RTX 4090D). To ensure transparency, we report all results alongside their corresponding control group baselines, presenting our findings as relative improvements over these baselines.

\begin{table}[tp]
  \centering
  \caption{Experimental results of GSC. We employe YOLO11n, YOLOv8 and YOLOv5 as our baseline models and conducted comprehensive retraining experiments across varying values of the smooth factor $\alpha$, respectively. We compare two important performance metrics for object detection, mAP@50-95 and mAP@50-95, with the best-performing configuration highlighted in bold.}
  \begin{tabular}{ccc}
    \hline
    Method & $\text{mAP}_{\text{test}}@\text{50--95}(\%)$ & $\text{mAP}_{\text{test}}@\text{50}(\%)$ \\
    \hline
    YOLO11n & 61.945 & 81.641 \\
    YOLO11n + GSC & \textbf{62.496} (\textbf{0.889\%$\uparrow$}) & \textbf{82.007} (\textbf{0.448\%$\uparrow$})\\ 
    YOLOv8 & 60.689 & 80.776 \\
    YOLOv8 + GSC &\textbf{60.897} (\textbf{0.343\%$\uparrow$})  &\textbf{81.390} (\textbf{0.760\%$\uparrow$})\\
    YOLOv5 &58.708& 79.694 \\
    YOLOv5 + GSC & 58.670(-0.006\%$\downarrow$) & \textbf{79.698} (\textbf{0.005\%$\uparrow$})  \\
    \hline
  \end{tabular}
  \label{tab:constant_result}
\end{table}

\subsection{Experimental Results of GSC}
\subsubsection{Comparison of Detection Accuracy}
As detailed in Section \ref{smooth constant}, the constant-scaled SBB strategy aims to find an average error value ($\varepsilon_{v}$) that brings most bounding boxes closer to the ideal bounding box label. By adopting this approach, we reformulate the challenging dataset error correction problem as a more tractable hyperparameter optimization task focused on determining the optimal smooth factor $\alpha$. We employ YOLO11n, YOLOv8 and YOLOv5 as our baseline models and conducted comprehensive retraining experiments across varying values of the smooth factor $\alpha$, respectively. We compare two important performance metrics for object detection, mAP@50-95 and mAP@50-95, the best results are shown in Table~\ref{tab:constant_result}. The results demonstrate that our SBB strategy achieves higher detection accuracy in most cases. Specifically, compared to the baseline methods YOLO11n, YOLOv8, and YOLOv5, mAP@50-95 improved by 0.889\%, 0.343\%, and -0.006\%, respectively, while mAP@50 improved by 0.448\%, 0.760\%, and 0.005\%, respectively.

\begin{figure}[tp]
    \centering
    \subfloat[Variation in mAP@50-95 scores with respect to different smooth factor values]{
        \begin{minipage}[b]{0.48\textwidth}
            \centering
            \includegraphics[width=\linewidth]{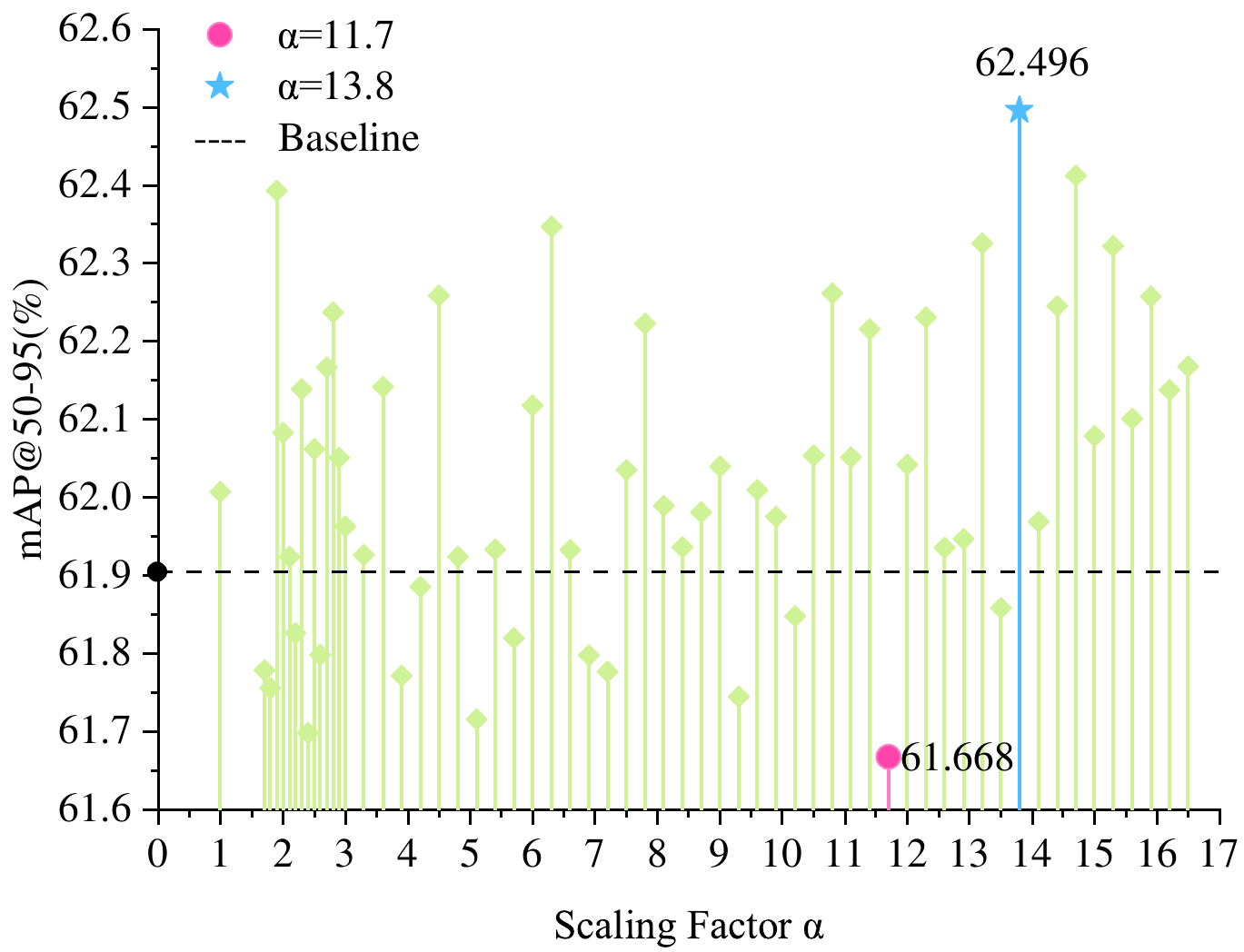}
            \label{fig:constant scaling1}
        \end{minipage}
    }
    \hfill
    \subfloat[Relative performance gains (\%) in mAP@50-95 compared to the baseline model over the range of smooth factors.]{
        \begin{minipage}[b]{0.48\textwidth}
            \centering
            \includegraphics[width=\linewidth]{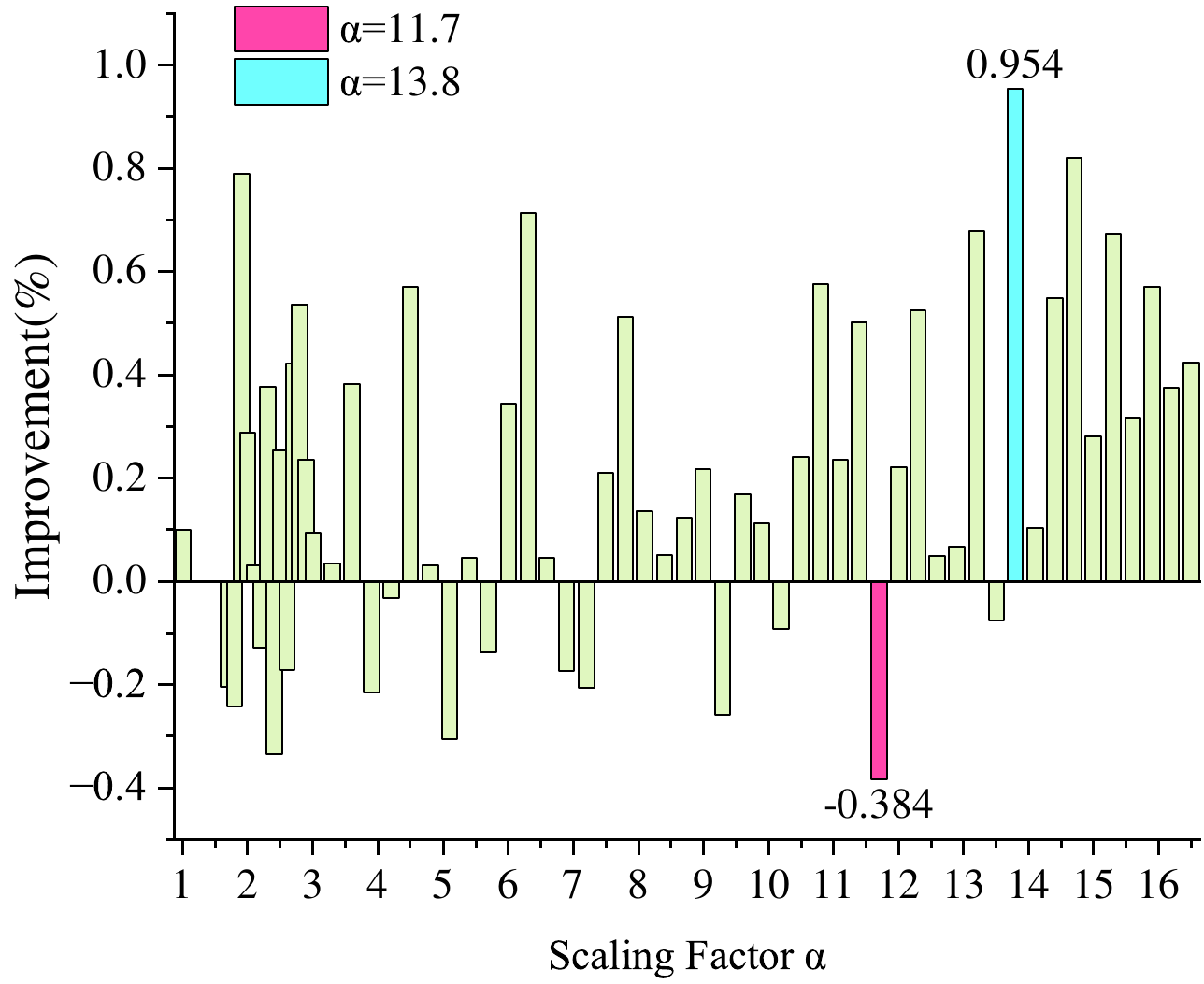}
            \label{fig:constant scaling2}
        \end{minipage}
    }
    \caption{Variation trends of both mAP@50-95 scores and their corresponding percentage improvements across different smooth factor values.}
    \label{fig:constant}
    \vspace{-0.2cm}	
\end{figure}

To provide a more intuitive demonstration of how the smooth factor affects detection accuracy, Figure~\ref{fig:constant} shows the variation trends of both mAP@50-95 scores (subfigure a) and their corresponding percentage improvements (subfigure b) across different smooth factor values, with YOLO11n as the baseline method.
The search range for $\alpha$ spans from 0 to 16.5. 
As presented in Figure~\ref{fig:constant}, an optimal smooth factor value can positively impact detection results, while an inappropriate value can have a negative effect. 
Specifically, when the smooth factor $\alpha = \text{13.80}$, the retrained model achieves an mAP@50-95 value of 0.6249 on the test set, reflecting a 0.5871\% improvement over the baseline which does not use the smoothing operation. 
Therefore, our proposed SBB strategy can effectively alleviate the impact of dataset errors and improve detection accuracy. 

\begin{figure}[tp]
    \centering
    \subfloat[Distribution of detection accuracy (mAP@50-95) across all 20 categories]{
        \begin{minipage}[b]{0.48\textwidth}
            \centering
            \includegraphics[width=\linewidth]{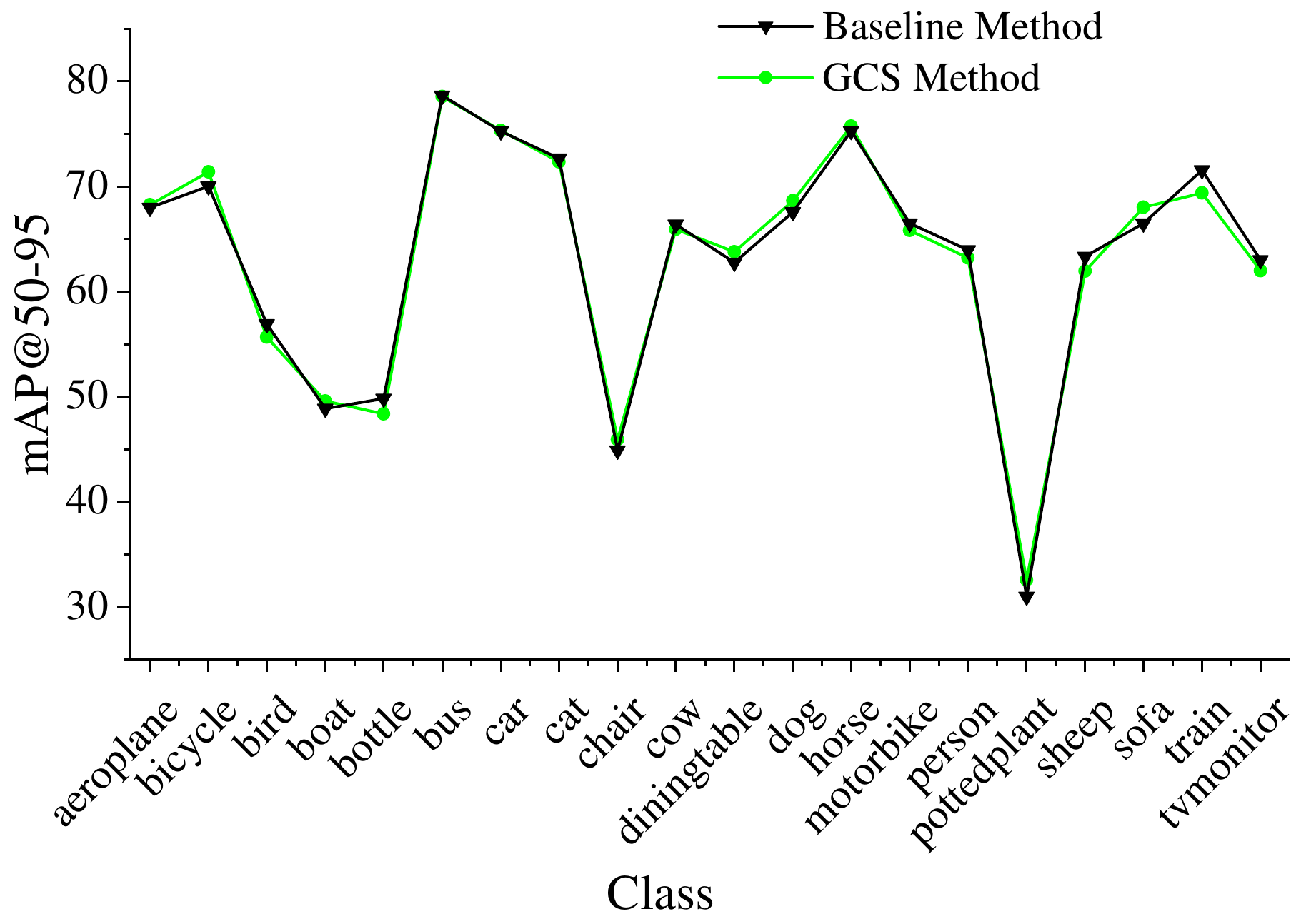}
            \label{fig:compare_class1}
        \end{minipage}
    }
    \hfill
    \subfloat[Relative improvement in detection accuracy achieved by our proposed approach compared to the baseline]{
        \begin{minipage}[b]{0.48\textwidth}
            \centering
            \includegraphics[width=\linewidth]{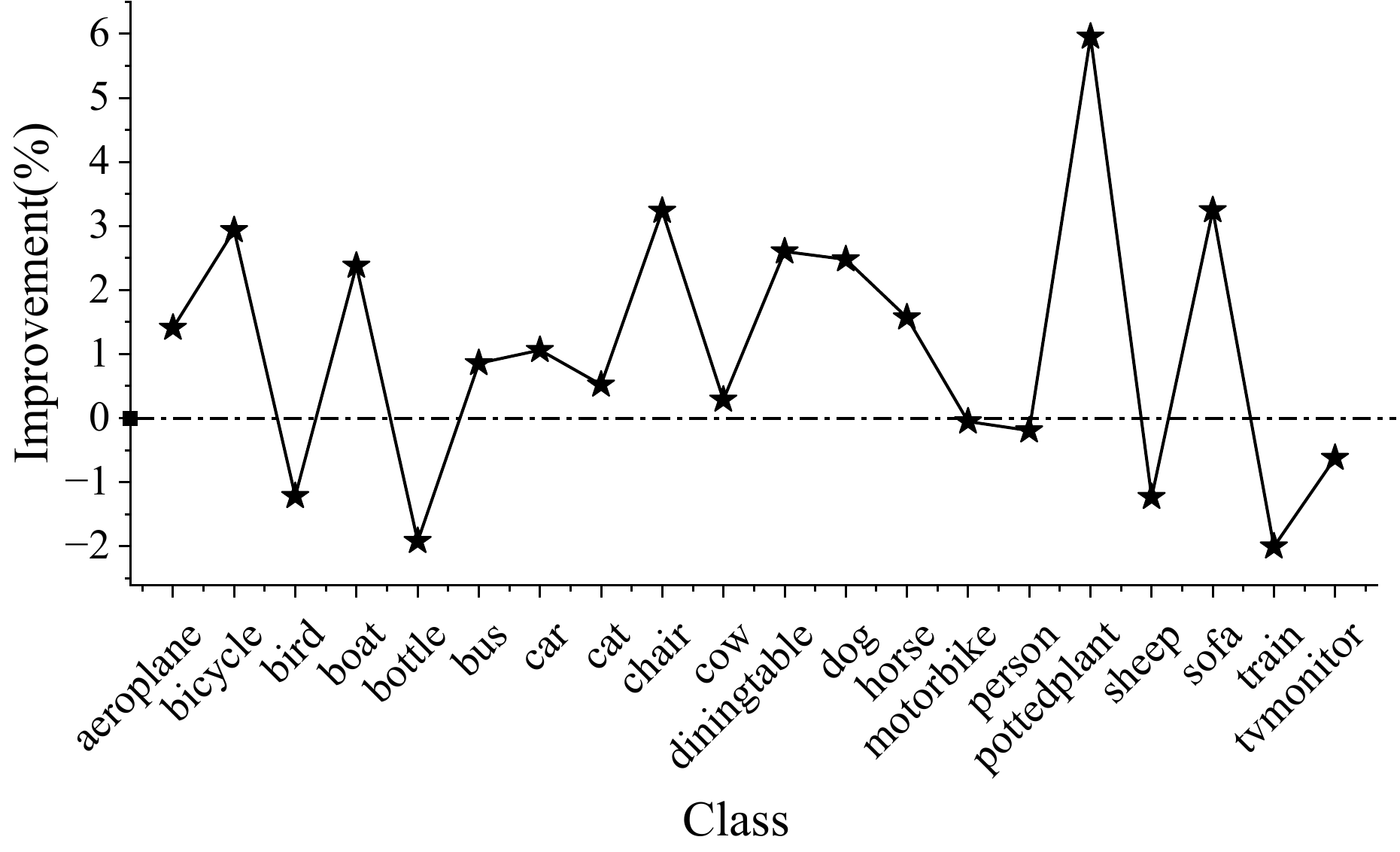}
            \label{fig:compare_class2}
        \end{minipage}
    }
    \caption{Comparison of detection accuracy across all 20 classes in the PASCAL VOC dataset, along with the corresponding performance gains relative to the YOLO11n baseline, using a smooth factor of $\alpha = \text{13.80}$.}
    \label{fig_all_class}
    \vspace{-0.2cm}	
\end{figure}

Furthermore, we conducted a comprehensive per-class analysis of detection accuracy and performance gains relative to the baseline, with the complete results visualized in Figure~\ref{fig_all_class}. Specifically, subfigure (a) presents the distribution of detection accuracy across all 20 categories in the PASCAL VOC dataset, where the green and black curves respectively depict the performance of our method GSC (at $\alpha = \text{13.80}$) and the YOLO11n baseline. Subfigure (b) quantifies the relative improvement in detection accuracy achieved by our proposed approach compared to the baseline. As shown in Figure~\ref{fig_all_class}, although our SBB strategy, which uses global scaling to estimate a single bounding box error to approximate annotation errors across the entire dataset, primarily focuses on this approximation, it still results in significant improvements in detection accuracy for most object categories. In contrast, while some categories (e.g., bird, bottle, sheep) show a slight reduction in classification accuracy, the extent of this degradation is much smaller than the gains observed in other categories. 

\begin{figure}[t]
    \centering
    \includegraphics[width=1\linewidth]{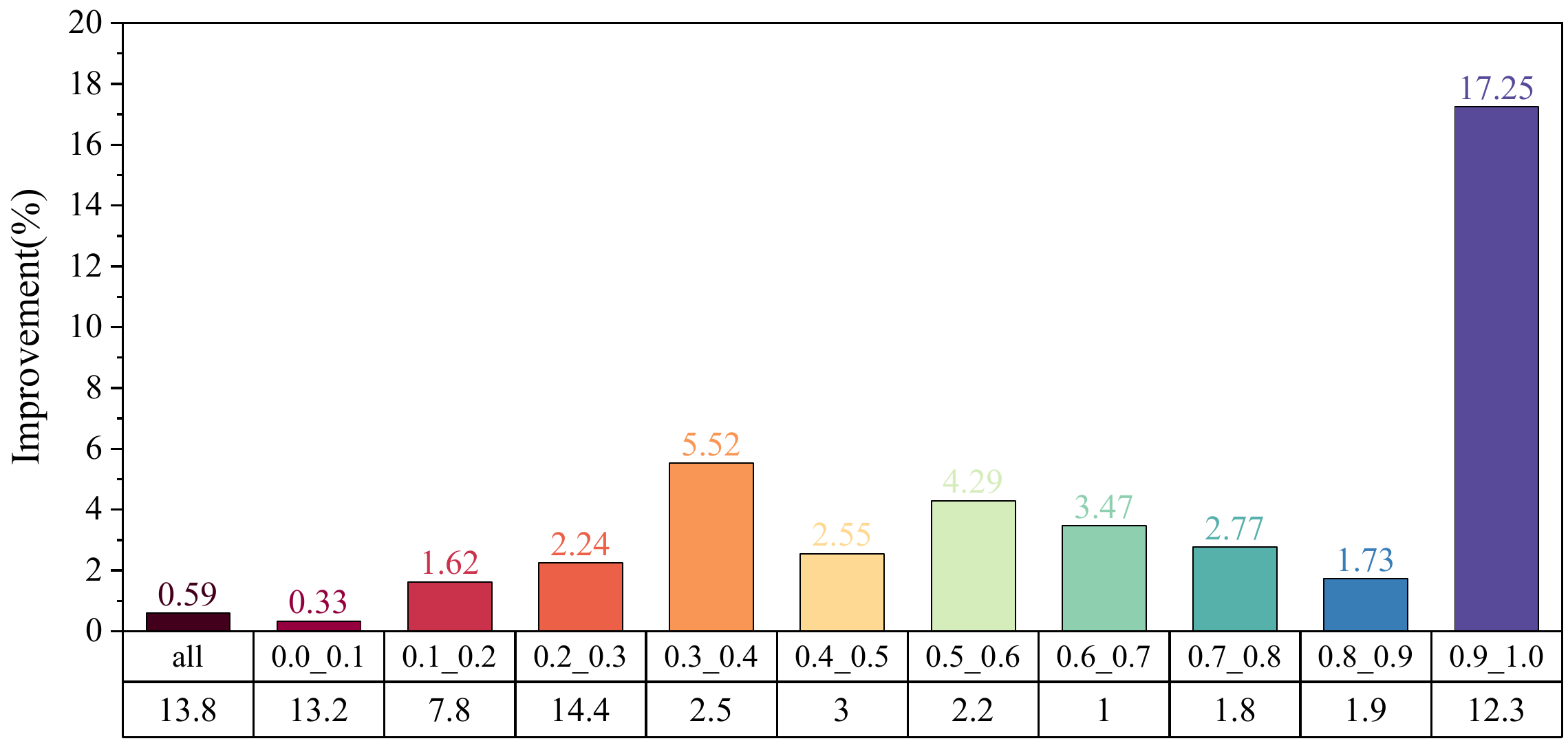}
    \caption{Comprehensive statistical analysis of objects across different 10 intervals during the smooth factor optimization process. The numbers in the bottom row represent the smooth factors corresponding to the interval where the objects achieved optimal detection performance (without considering the overall detection performance). The Y-axis shows the relative improvement in mAP@50-95 achieved compared to the baseline.}
    \label{fig_constant_interval}
    	\vspace{-0.3cm}	
\end{figure}

This phenomenon can be attributed to the imbalanced distribution within the dataset, where small objects significantly outnumber large ones. This pattern is consistently evident both in the overall dataset composition (see Figure~\ref{fig_datasets}) and across individual object categories (see Figure~\ref{fig:class_distribution}). When annotating objects of varying sizes, labelers inevitably introduce scale-dependent errors. Using a single estimated error value for bounding box correction across all scales leads to compromised performance. For example, while optimizing for small objects, the correction fails to account for objects of different sizes, ultimately reducing detection accuracy. 
To validate the reliability of this conclusion, we conducted a comprehensive statistical analysis of objects across different intervals during the smooth factor optimization process. 
Specifically, a local optimization strategy was used to independently determine the most suitable smooth factors for each interval, without considering the overall detection performance. The results are shown in Figure~\ref{fig_constant_interval}, which reveals notable variations in the optimal parameters across different intervals. For example, objects within the relative size ranges of [0, 0.1] and [0.9, 1] achieve their optimal detection performance at smooth factors of $\alpha=\text{13.2}$ and $\alpha=\text{12.3}$, respectively. Although objects in the [0.9, 1] size range show significantly greater detection accuracy improvement than those in the [0, 0.1] range, the globally optimized smooth factor favors the [0, 0.1] range due to dataset imbalance, where small objects substantially outnumber large ones. Therefore, it is essential to treat objects in different intervals separately and optimize a distinct smooth factor for each.

To sum up, although our smooth bounding
box strategy via constant scaling may not fully accommodate objects of all sizes simultaneously, it effectively improves overall detection accuracy on a global scale. Given the thousands of training images, estimating a single average labeling error value is highly cost-effective, making it a practical approach.

\subsubsection{Delay the Onset of Overfitting}
As established in Section~\ref{nrf}, our SBB formulation enhances model robustness against annotation errors by systematically mitigating the disproportionate loss values induced by bounding box inaccuracies. 
To prove the correctness of this conclusion, we conduct a performance comparison from the quantitative and qualitative perspectives. 

\begin{figure}[t]
    \centering
    \includegraphics[width=1\linewidth]{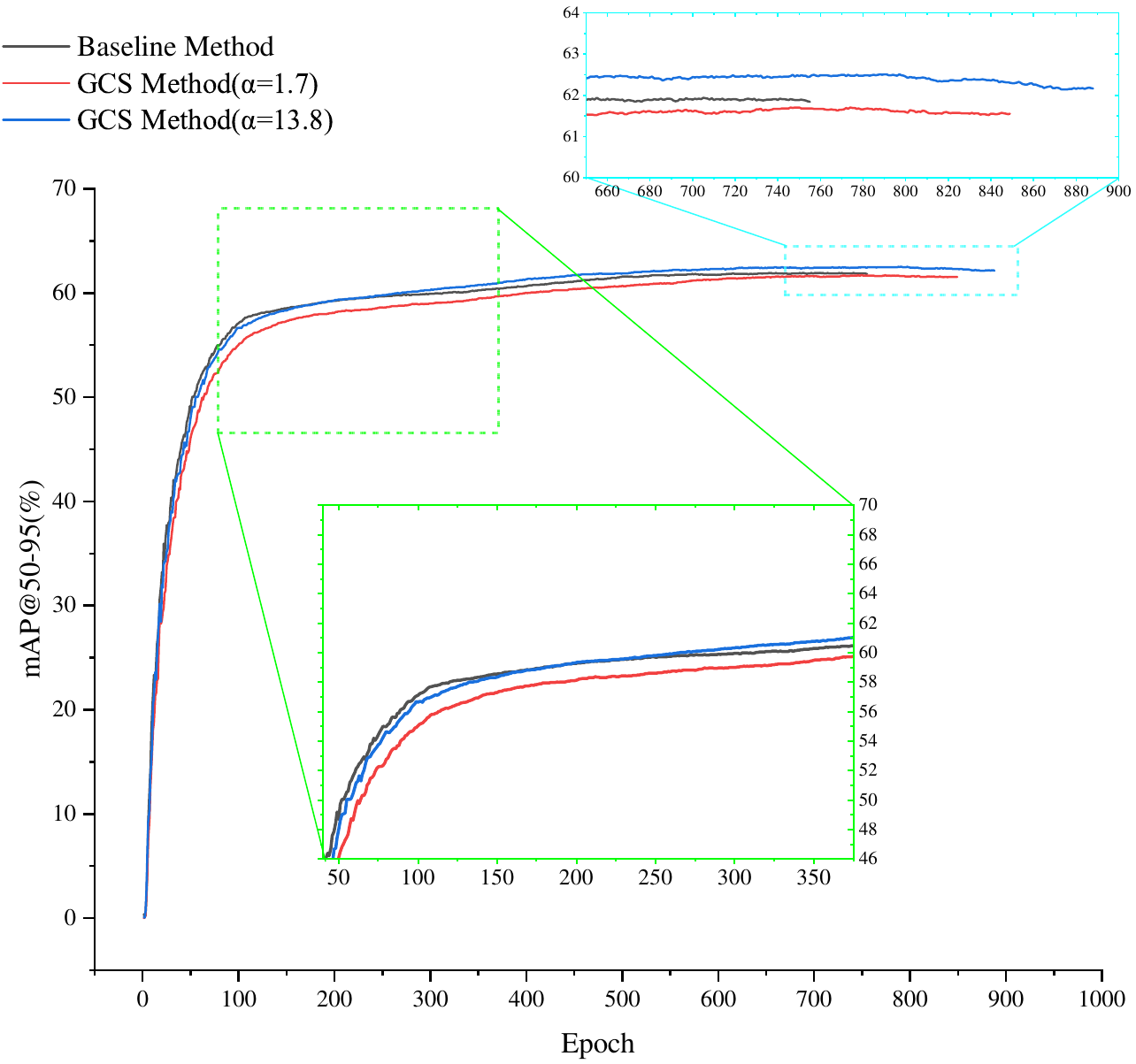}
    \caption{mAP@50-95 curves on the validation set during the model training process are shown. The black curve represents the baseline YOLO11n, while the blue curve represents the SBB method with $\alpha=1.70$ (the worst smooth factor which exhibited negative effects), and the SBB method with $\alpha=13.80$ (the best smooth factor which archived positive effects). The process of the blue curve surpassing the black curve is clearly visible in the green magnifying window.}
    \label{fig_epoch}
    \vspace{-0.3cm}	
\end{figure}

\begin{figure}[t]
    \centering
    \includegraphics[width=1\linewidth]{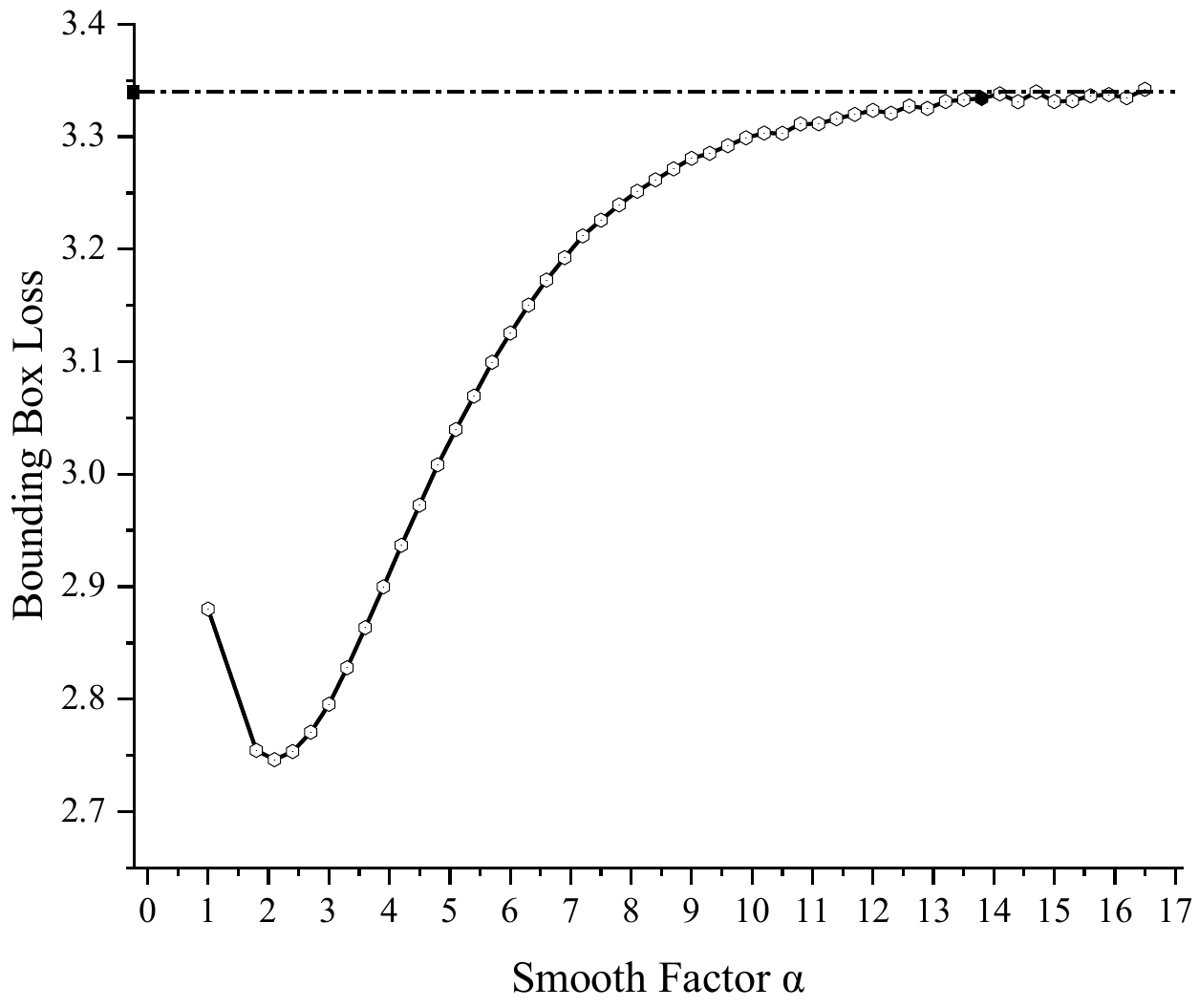}
    \caption{Variation of our method's first-epoch bounding box loss across different $\alpha$ values. The dashed line indicates the baseline's first-epoch loss.}
    \label{fig:boxloss}
\end{figure}

From a qualitative perspective, we plotted the mAP@50-95 curves on the validation set during the model training process, and the results are shown in Figure~\ref{fig_epoch}. 
We selected three experimental groups: (1) the baseline YOLO11n, (2) the SBB group with $\alpha=\text{1.70}$, the worst smooth factor for constant sclaing operation. (3) the SBB group with $\alpha=\text{13.80}$, the best smooth factor for constant sclaing operation. The $\alpha=\text{1.70}$ group exhibited negative effects compared to the baseline, while the $\alpha=\text{13.80}$ group showed positive effects.
Figure~\ref{fig_epoch} demonstrates that while the smooth factor's effect may vary in direction (positive/negative), it consistently delays overfitting onset. 
Specifically, the baseline consistently outperformed the SBB groups during the early stages of training. However, when the number of epochs reached between 150 and 250, their performance became comparable. Beyond this point, the group with $\alpha=\text{13.80}$ surpassed the baseline.
In the final stages of training, the group with $\alpha=\text{1.70}$ delayed the onset of overfitting by approximately 100 epochs, while the group with $\alpha=\text{13.80}$ delayed overfitting onset by around 150 epochs. Additionly, since our method directly optimizes the bounding box loss and inherently introduces loss smoothing, we further analyze the initial training behavior by evaluating the bounding box loss at epoch=1 across different values of $\alpha$. The results are presented in Figure~\ref{fig:boxloss}. The figure shows that when an appropriate smooth factor is selected, such as $\alpha=\text{13.80}$, the initial bounding box error of the model training is not significantly reduced. Therefore, even with our SBB strategy, the model training efficiency will not be substantially compromised, as demonstrated in Figure~\ref{fig_epoch}. 

\begin{table}[tp]
\centering
\caption{NRF values evaluated across multiple metrics, including mAP@50-95, training loss, and validation loss throughout the model training process.}
\begin{tabular}{cccc} \hline 
\multirow{2}{*}{Method} & \multicolumn{3}{c}{NRF}\\ 
\cline{2-4}
 &$\text{mAP@}\text{50-95}$&$loss_\text{train}$&$loss_{\text{val}}$\\ \hline
Baseline(YOLO11n) & 0.002869&0.051160&0.0031870\\ 
Our ($\alpha=1.70$)& 0.002503$\downarrow$ &0.006663$\downarrow$ &0.003200$\uparrow$\\ 
Our ($\alpha=13.80$)& \textbf{0.002482}$\downarrow$&\textbf{0.005945}$\downarrow$&\textbf{0.002659}$\downarrow$\\ \hline
\end{tabular}
\label{tab:nfa_results}
\end{table}

From a quantitative perspective, we use the NRF metric to evaluate the smoothness of the model training process.
Specifically, NRF is employed to evaluate both the stability of the mAP@50-95 metric and the loss values (including training loss and validation loss) throughout the model training process. The results are shown in Table~\ref{tab:nfa_results}. It can be observed that the baseline achieves a higher NRF value (0.002869) for mAP@50-95 compared to our SBB strategy with global scaling ($\alpha=\text{1.7}$ and $\alpha=\text{13.8}$), which means a more unstable training process. 
In terms of the NRF values of the losses, we observe that our method with the best parameter $\alpha=13.8$ outperforms the baseline on both the training loss and validation loss. However, although the NRF of the control group ($\alpha=1.7$) shows an improvement in training loss compared to the baseline, it does not offer an advantage in validation loss. This explains why the detection performance of our method with this $\alpha$ value is not as good as the baseline. 

To sum up, our proposed method achieves the best NRF values across various metrics when the smooth factor is set to an appropriate value.
Therefore, we have demonstrated that our method enhances model robustness against annotation errors from both quantitative and qualitative perspectives.

\subsection{Experimental Results of ISC}
\subsubsection{Parameters Selection}
Since object detection networks typically employ an end-to-end training strategy that combines classification losses and anchor box regression losses for all objects before performing unified gradient backpropagation, the annotation errors of different objects can influence one another. Therefore, although we have analyzed the optimal smooth factors for objects of different relative sizes across various intervals in the experiments of GSC, simply applying the parameters listed in Figure~\ref{fig_constant_interval} to the interval scaling correction strategy cannot yield good detection performance for the models. However, simultaneously searching for a smooth factor for each interval creates a high-dimensional optimization problem, posing significant computational challenges. For example, when objects are divided into $m$ intervals, each with $n$ candidate smoothing factors, the total search space grows exponentially to $n^m$. Therefore, we employed a fine-tuning approach that prioritizes processing the interval with the highest number of objects (i.e., the interval of [0, 0.1]), and then optimizes the smooth factors for other intervals through fine-tuning.

\begin{table*}[tp]
  \centering
  \caption{$\text{mAP}_\text{val}\text{@50-95}$ results for the three training methods (TFS, FT, RT) with smooth factors ranging from 1 to 9 on dataset PASCAL VOC. The $\text{mAP}_\text{val}\text{@50-95}$ for the baseline YOLO11n is 61.911.}
  
    \begin{tabular}{cccccccccc}
    \toprule
\multirow{2}{*}{Method} &\multicolumn{9}{c}{Smooth Factor ($\alpha$)}\\ \cline{2-10}
 & 1 & 2& 3 & 4 & 5& 6& 7 & 8& 9 \\ \hline
TFS&  61.673   & 61.631   & 61.542   & 61.804   & 61.840   & 62.063   & 61.614   & \textbf{62.512} & 61.822  \\
FT &  \textbf{61.624} & 61.513   & 61.542   & 61.596   & 61.254   & 61.432   & 61.486   & 61.386   & 61.621  \\
RT & \textbf{62.371} & 62.137   & 62.285   & 61.998   & 62.287   & 62.185   & 62.266   & 62.220   & 62.252  \\
    \bottomrule
    \end{tabular}%
  \label{tab:trainmethod}%
\end{table*}%

To identify the optimal training approach for the interval scaling correction method, determine the initial model training starting point, and find the best smooth factor for the first interval, we explored three distinct methodologies: training from scratch (TFS), where the model is trained entirely from the beginning; fine-tuning (FT), which adjusts parameters based on pre-trained models; and resume training (RT), where training continues from a specific checkpoint.
Specifically, for TFS, we trained the model using nearly the same settings as the baseline, but incorporated our smoothing bounding box method for objects with relative sizes within the [0, 0.1] interval. 
For FT, we performed fine-tuning on objects within this same size range, based on the baseline model. In other words, we retrained these objects using nine smooth factors (ranging from 1 to 9), while leaving other objects unaffected by the smoothing operation. 
For RT, two starting points were used to resume model training: the TFS model with the smoothing method and the baseline model. The process is similar to FT, where early stopping controls the training, and the best weights are selected for further training.
The key difference between FT and TFS lies in checkpoint initialization. In FT, the model fully inherits all hyperparameters from the checkpoint, preserving the exact model state (including epoch count) at the moment training was stopped. 
  
\begin{table*}[tp]
  \centering
  \caption{$\text{mAP}_\text{val}\text{@50-95}$ results obtained by retraining and fine-tuning the models (include YOLO11n, YOLOv8, YOLOv5) using different smooth factors within the range [1, 9]. The results in the gray region (i.e., the interval [0, 0.1]) were obtained using the TFS training method, while the other results were obtained using the RT method. The data in the [0.1, 0.3] interval were fine-tuned based on the optimal results already achieved in [0, 0.1], and the interval [0.3, 0.5] was fine-tuned based on the optimal results in [0.1, 0.3]. The same approach was applied to the other intervals.}
    \begin{tabular}{ccccccccccc}
    \toprule
    \centering
    \multirow{2}{*}{Baseline} &\multirow{2}{*}{Interval} & \multicolumn{9}{c}{Smooth Factor ($\alpha$)}\\
    \cline{3-11}
    && 1& 2& 3& 4& 5& 6& 7 & 8& 9 \\
    \hline
     \multirow{5}{*}{\makecell{YOLO11n\\(61.911)}}&
     \cellcolor{gray}[0.0,0.1] & \cellcolor{gray}61.673  & \cellcolor{gray}61.631  & \cellcolor{gray}61.542  & \cellcolor{gray}61.804  & \cellcolor{gray}61.84 & \cellcolor{gray}62.063 & \cellcolor{gray}61.614 & \cellcolor{gray}\textbf{62.512}$\uparrow$ & \cellcolor{gray}61.822 \\
     &{[0.1,0.3]} &  62.746 &  62.718 &  62.806 &  62.573 &  62.778 &  62.673 &  62.649 &  \textbf{62.909}$\uparrow$ &  62.762 \\
    &{[0.3,0.5]} & 62.717  & 62.679  & 62.698  & 61.54   & 62.781  & 62.514  & 61.476  & 61.532  & 62.575 \\&
    {[0.5,0.7]} & 61.688  & 62.63   & 61.542  & 62.605  & 61.548  & 62.824  & 61.605  & 61.605  & 62.581 \\&
    {[0.7,1.0]} & 61.602  & 61.606  & 62.576  & 62.667  & 61.605  & 61.605  & 62.810   & 62.780   & 62.507 \\
    \hline
    \multirow{5}{*}{\makecell{YOLOv8\\(60.647)}}&\cellcolor{gray}[0.0,0.1] &  \cellcolor{gray}60.994 &  \cellcolor{gray}60.943 &  \cellcolor{gray}\textbf{61.020}$\uparrow$ &  \cellcolor{gray}60.653 &  \cellcolor{gray}60.664 & \cellcolor{gray}60.859 & \cellcolor{gray}60.944 &  \cellcolor{gray}60.463 &  \cellcolor{gray}60.594 \\&
    {[0.1,0.3]} &  61.258 &  61.144 &  61.241 &  \textbf{61.148}$\uparrow$ &  61.331 &  61.194 &  61.383 &  61.092 &  61.319 \\&
    {[0.3,0.5]} &  61.342 &  61.472 &  61.264 &  61.307 &\textbf{61.486}$\uparrow$ &  61.191 &  61.333 &  61.257 &  61.356 \\&
    {[0.5,0.7]} & 61.353 & 61.387 & 61.479 & 61.241  & 61.390 & 61.293 & 61.436 &  61.428 & 61.281 \\&
    {[0.7,1.0]} & 61.317 & 61.338 & 61.288 & 61.522 & 61.335 & \textbf{61.559$\uparrow$} & 61.391 & 61.391 & 61.391 \\
    \hline
    \multirow{5}{*}{\makecell{YOLOv5\\(58.675)}}&\cellcolor{gray}[0.0,0.1] & \cellcolor{gray}59.100 & \cellcolor{gray}59.258 & \cellcolor{gray}59.307 & \cellcolor{gray}59.066 & \cellcolor{gray}59.138 & \cellcolor{gray}\textbf{59.343}$\uparrow$ &\cellcolor{gray}59.069 & \cellcolor{gray}59.246 & \cellcolor{gray}59.211 \\&
    [0.1,0.3] & 59.464 & 58.155 & 59.440 & \textbf{59.599}$\uparrow$ & 58.1337 & 58.210 & 58.136 & 58.158 & 59.393 \\&
    [0.3,0.5] & 59.533 & \textbf{59.851}$\uparrow$ & 58.468 & 59.651 & 58.354 & 58.425 & 58.509 & 58.435 & 58.372 \\&
    [0.5,0.7] & 58.603 & 58.625 & \textbf{60.034}$\uparrow$ & 59.842 & 59.790 & 58.603 & 58.403 & 58.522 & 59.820 \\&
    [0.7,1.0] & 58.505 & 59.772 & 59.677 & 58.519 & 58.519 & 59.843 & 58.519 & 58.519 & 58.519 \\
    \bottomrule
    \end{tabular}
  \label{tab:interval}
\end{table*}

The $\text{mAP}\text{@50-95}$ values on the PASCAL VOC validation set. obtained by three training methods across various smooth factors, are summarized in Table~\ref{tab:trainmethod}. Table~\ref{tab:trainmethod} presents two key findings: First, the TFS method significantly outperforms both FT and RT in terms of $\text{mAP}\text{@50-95}$ when the smooth factor is appropriately chosen. Second, the RT method demonstrates greater stability, consistently maintaining positive performance gains across all $\alpha$ values in the range [1, 9]. While RT's accuracy does not match that of TFS, it remains competitive. 
Based on these experimental observations, we adopted the following optimization strategy: First, the TFS method is used to fine-tuning the objects within the [0, 0.1] interval to identify the optimal value of $\alpha$. Next, leveraging the training results obtained with this optimal $\alpha$ in the [0, 0.1] interval, we iteratively apply the RT method to determine the best $\alpha$ values for the remaining intervals, progressively refining the model from smaller to larger intervals, ultimately yielding the final optimized model. 

We retrained and fine-tuned the model on the PASCAL VOC validation set using different smooth factor values. The resulting detection performance is presented in Table~\ref{tab:interval}. Specifically, considering the non-uniform distribution of the dataset, we divided objects into five uneven intervals based on relative size: [0, 0.1], [0.1, 0.3], [0.3, 0.5], [0.5, 0.7], and [0.7, 1].
For the smooth factor parameter, we conducted a comprehensive grid search across all integer values in the discrete set $\{\text{1}, \text{2}, ..., \text{9}\}$ to evaluate their respective impacts. 
As shown in Table~\ref{tab:interval}, our model training approach can effectively improve the detection accuracy and efficiently identify the optimal value of the smooth factors for objects of different size intervals. 
If we directly search for the optimal parameter combination for objects in all five intervals simultaneously, the dimensionality of the search space would be $\text{9}^\text{5}=\text{59049}$, which is significantly larger than the current $\text{5}*\text{9}=\text{45}$.
Furthermore, refining either the interval partitioning granularity or the smooth factor search range would induce an exponential increase in dimensionality for the direct search method, while our method only grows in a multiplicative manner.

Based on the aforementioned training approach, Table~\ref{intervalfactor} presents the optimal smooth factor values for different object intervals, as obtained by the three baseline methods on the PASCAL VOC dataset. A value of 0 indicates that no smoothing operation was applied.
Due to the varying image features focused on by different models, the resulting loss differs, and as a result, the corresponding smooth factor values for each interval also vary. 
The observations from Table~\ref{intervalfactor} and Table~\ref{tab:interval} reveal a deeper issue: In intervals with sparse data distribution, the limited computational resources that constrain the search space will make it challenging for our method to achieve a balanced solution across all objects. Additionally, the end-to-end training paradigm hinders the full capture of the causal mechanism. Thus, in practical applications with ample computational resources, we recommend using denser objects interval partitions for more precise control.

\begin{table}[tp]
  \centering
  \caption{The optimal smooth factor value for each interval on dataset PASCAL VOC. A value of 0 presents that the objects in the corresponding interval do not use the SBB operation.}
    \begin{tabular}{cccccc}
    \toprule
    \centering
    \multirow{2}{*}{Baseline} &\multicolumn{5}{c}{Interval} \\
    \cline{2-6}
    & [0, 0.1]&[0.1, 0.3]&[0.3, 0.5]&[0.5, 0.7]&[0.7, 1] \\
    \hline
    YOLO11n& 8 &8&0&0&0 \\
    YOLOv8n& 3 &4&5&0&4 \\
    YOLOv5n& 6 &4&2&5&0 \\    
    \bottomrule
    \end{tabular}
  \label{intervalfactor}
\end{table}

\begin{table}[tp]
  \centering
  \caption{Experimental results of ISC on the PASCAL VOC and COCO datasets. We employ YOLO11n, YOLOv8, and YOLOv5 as our baseline models and conducted comprehensive retraining experiments across various values of the smooth factor. We compare two important performance metrics for object detection, mAP@50 and mAP@50-95, with the best-performing ones highlighted in bold.}
  \begin{tabular}{cccc}
    \hline
    Dataset&Method & $\text{mAP}_{\text{test}}@\text{50--95}(\%)$ & $\text{mAP}_{\text{test}}@\text{50}(\%)$ \\
    \hline
    \multirow{6}{*}{VOC}&YOLO11n & 61.945 &  81.641\\
    &YOLO11n + ISC & \textbf{62.872}(\textbf{2.415\%})$\uparrow$ & \textbf{82.334} (\textbf{1.039\%})$\uparrow$\\ 
    &YOLOv8 & 60.689 &  80.776\\
   & YOLOv8 + ISC & \textbf{61.519}(\textbf{2.254\%})$\uparrow$& \textbf{81.390} (\textbf{0.940\%})$\uparrow$  \\
    &YOLOv5 & 58.707&79.694  \\
    &YOLOv5 + ISC & \textbf{60.072}(\textbf{3.690\%})$\uparrow$& \textbf{80.899} (\textbf{1.897\%})$\uparrow$  \\
    \hline
    \multirow{6}{*}{COCO}&YOLO11n & 39.9 &  55.9\\
    &YOLO11n + ISC & \textbf{40.0}(\textbf{0.25\%})$\uparrow$ & \textbf{56.0} (\textbf{0.17\%})$\uparrow$\\ 
    &YOLOv8 & 38.0 &  53.7\\
    &YOLOv8 + ISC & \textbf{38.5}(\textbf{1.31\%})$\uparrow$& \textbf{54.1} (\textbf{0.74\%})$\uparrow$  \\
    &YOLOv5 & 35.5&51.1  \\
    &YOLOv5 + ISC & \textbf{35.7}(\textbf{0.56\%})$\uparrow$& \textbf{51.2} (\textbf{0.19\%})$\uparrow$  \\
    \hline
  \end{tabular}
  \label{tab:interval_result}
\end{table}

\subsubsection{Comparison of Detection Accuracy}
Table~\ref{tab:interval_result} presents the experimental results of our ISC method (with optimal smoothing factors) against the baselines. Similar to the experiment in the GSC method, we also compared the metrics of mAP@50-95 and mAP@50-95. The results indicate that the proposed SBB strategy yields superior detection accuracy. Specifically, it outperforms the baseline models (YOLOv5, YOLOv8, and YOLO11n), achieving improvements of [3.690\%, 2.254\%, 2.415\%] in mAP@50–95 and [1.897\%, 0.940\%, 1.039\%] in mAP@50, respectively. Furthermore, the effectiveness of our method was further validated by experimental results from testing the ISC method on the COCO dataset.
In comparison between GSC and ISC, results from Tables~\ref{tab:constant_result} and~\ref{tab:interval_result} indicate that the ISC method achieves significant improvement over GSC. This demonstrates the practical effectiveness of our proposed SBB strategy.

\begin{figure}[t]
    \centering
    \includegraphics[width=1\linewidth]{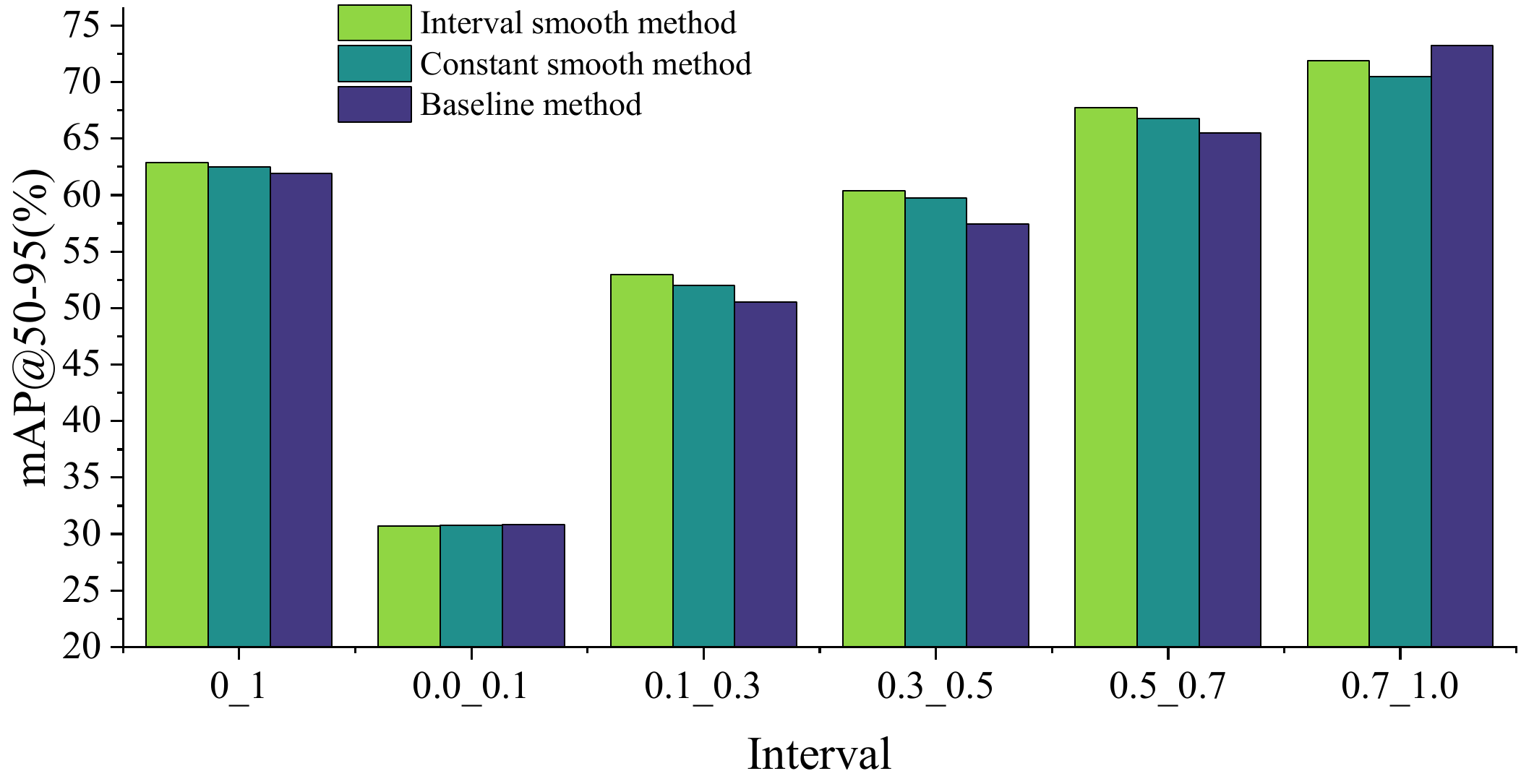}
    \caption{The detection accuracy (mAP@50-95) on the PASCAL VOC dataset across different intervals is shown below. The dark blue, green, and light blue bars correspond to the results of the baseline, YOLOv11n+ISC (with parameters from Table~\ref{intervalfactor}), and the GSC method, respectively.}
    \label{fig:Interval5}
    \vspace{-0.3cm}	
\end{figure}

Furthermore, we analyze the performance of our method from a partitioning perspective. Specifically, objects are divided into five intervals ([0,0.1], [0.1,0.3], [0.3,0.5], [0.5,0.7], [0.7,1]) based on their relative sizes. The detection accuracy for objects within each interval was evaluated separately. The results for the PASCAL VOC dataset are presented in Figure~\ref{fig:Interval5}. As shown in Figure~\ref{fig:Interval5}, while maintaining the detection accuracy for small objects that constitute the majority of the dataset, the ISC method achieves significant improvement over GSC in detecting objects across the other four intervals. Therefore, treat the objects in different intervals separately and optimize a distinct smooth factor for each is meaningful.

\section{Conclusion}
\label{conclu}
This paper proposed a novel SBB strategy to mitigate detection accuracy degradation caused by annotation inaccuracies. We developed two implementation approaches: GSC for dataset-wide error estimation, and ISC for size-adaptive refinement. Both methods transform error correction into parameter optimization problems without increasing inference complexity.
Experimental results demonstrate that our approach consistently improves detection accuracy across multiple baselines on PASCAL VOC and COCO datasets. The method also enhances training stability, as verified by our proposed Normalized Residual Fluctuation metric.
Future work may explore automated smooth factor learning and extension to other vision tasks affected by label noise. The proposed strategy offers an effective solution for developing more robust object detection systems.



 
%


\bibliographystyle{IEEEtran} 
\bibliography{refer}

\end{document}